\documentclass[letterpaper]{article} % DO NOT CHANGE THIS
\usepackage{aaai2027}  % DO NOT CHANGE THIS
\usepackage[hyphens]{url}  % DO NOT CHANGE THIS
\usepackage{graphicx} % DO NOT CHANGE THIS
\usepackage{natbib}  % DO NOT CHANGE THIS AND DO NOT ADD ANY OPTIONS TO IT
\usepackage{caption} % DO NOT CHANGE THIS AND DO NOT ADD ANY OPTIONS TO IT
\usepackage{algorithm}
\usepackage{algorithmic}
\usepackage{tabularx}
\usepackage{xcolor}
\usepackage{tcolorbox}
\usepackage{amsmath}
\usepackage{amssymb}
\usepackage{wasysym}
\usepackage{multirow}
\usepackage{xspace}

\usepackage{newfloat}
\usepackage{listings}
\DeclareCaptionStyle{ruled}{labelfont=normalfont,labelsep=colon,strut=off} % DO NOT CHANGE THIS
\floatstyle{ruled}
\newfloat{listing}{tb}{lst}{}
\floatname{listing}{Listing}

\usepackage{booktabs}

\newcommand{\etal}{\textit{et al.\@ }}
\title{TTSD-FAR: Test-Time Self-Distillation with Fisher-Anchored Restoration for Adaptation of LVLMs under Missing Modalities}

\title{TTSD-FAR: Test-Time Self-Distillation with Fisher-Anchored Restoration of LVLMs for Missing-Modality Emotion Recognition}

\title{TTSD-FAR: Test-Time Self-Distillation with Fisher-Anchored Restoration for Missing-Modality Emotion Recognition in LVLMs}

\author{
    Written by AAAI Press Staff\textsuperscript{\rm 1}\thanks{With help from the AAAI Publications Committee.}\\
    AAAI Style Contributions by Peter Patel Schneider,
    Sunil Issar,\\
    J. Scott Penberthy,
    George Ferguson,
    Hans Guesgen,
    Francisco Cruz\equalcontrib\corresponding,
    Marc Pujol-Gonzalez\equalcontrib\corresponding
}
\affiliations{
    \textsuperscript{\rm 1}Association for the Advancement of Artificial Intelligence\\
    1101 Pennsylvania Ave, NW Suite 300\\
    Washington, DC 20004 USA\\
    proceedings-questions@aaai.org
}

\title{TTSD-FAR: Test-Time Self-Distillation with Fisher-Anchored Restoration for Missing-Modality Emotion Recognition in LVLMs}
\author {
    Muhammad Haseeb Aslam \textsuperscript{\rm 1},
    Alessandro Koerich \textsuperscript{\rm 1},
    Marco Pedersoli \textsuperscript{\rm 1},
    Ali Etemad \textsuperscript{\rm 2},
    Eric Granger\textsuperscript{\rm 1}
}
\affiliations {
    \textsuperscript{\rm 1}LIVIA, ETS Montreal, Canada  \\
    \textsuperscript{\rm 2}Aiim Lab, Queen's University, Canada. \\
    muhammad-haseeb.aslam.1@ens.ettsmtl.ca
}
\begin{document}

\maketitle

\begin{abstract}
Large video-language models (LVLMs) have shown remarkable performance on multimodal tasks like multimodal emotion recognition (ER) in-the-wild. ER is inherently multimodal, requiring a joint understanding of facial expressions, vocalizations, language, biosignals, and gestures. However, real-world deployment remains challenging: modalities may be missing or noisy at test time. Partial observations can be viewed as a distribution shift relative to the complete-modality distribution. State-of-the-art TTA methods based on entropy minimization or perplexity reduction do not transfer to autoregressive LVLMs, while retrieval-augmented generation (RAG) degrades when the observed modality is weak. 
%Since no ground-truth supervision exists, test-time adaptation (TTA) must proceed continuously, risking degradation once the model drifts from a reliable solution. 
Because no ground-truth supervision exists to verify individual updates, adaptation across this stream risks accumulating drift and degrading once the model departs from a reliable solution.
An effective solution must therefore adapt to arbitrary missing-modality patterns and remain effective during continual adaptation. 
%%%%
We address both jointly with \textbf{Test-Time Self-Distillation (TTSD)}, a parameter-efficient framework in which a frozen teacher, trained on complete modalities, guides an adaptive low-rank student via self-distillation, updating only a negligible number of parameters. Stability is built into this same loop through \textbf{Fisher-Anchored Restoration (FAR)}, which monitors Fisher information stability to detect convergence versus drift and restores the student toward the teacher's anchor when distributional shifts are identified.
%%%%
Our experiments\footnote{Our code is included in the supplementary materials and will be made public.} on MELD, DFEW, and BAH under 0\%--50\% missing modalities show that this unified adaptation-restoration design consistently outperforms entropy-based adaptation, RAG, and perplexity-based generation over long adaptation horizons, where baselines without restoration progressively degrade while TTSD-FAR remains consistent.
\end{abstract}

\section{Introduction}
\label{sec:intro}
%%%%%%%%%%%%%%%%%%%%%%% 
The rapid advancement of LVLMs \cite{videollava2023, videochatgpt2023, li2023videochat} has revolutionized multimodal learning tasks, achieving unprecedented performance.  These models leverage transformer architectures with billions of parameters, pre-trained on massive video-text corpora, to develop rich cross-modal representations \cite{radford2021clip}. LVLMs are promising for video-based multimodal ER (MER), where systems typically capture facial, vocal, and language cues to address subtle (or compound) expressions and high inter-subject variability \cite{huang2025emotionqwenunifiedframeworkemotion, cheng2024emotionllamamultimodalemotionrecognition}.  
% on benchmarks spanning video captioning, visual question answering, and even predictive tasks like emotion recognition. 
Recent work shows that LVLMs can capture fine-grained emotional and health states and contextual relationships that traditional discriminative models fail to detect \cite{ge2024videoemotionopenvocabularyrecognition}. However, their deployment in real-world applications confronts a fundamental challenge that has received surprisingly little attention: missing modality at test time. 

In practical video-based MER scenarios, modality absence is ubiquitous and inevitable. Inputs may suffer from audio corruption in noisy environments, transcription services may fail due to poor audio quality or privacy constraints, and sensor malfunctions can eliminate entire modalities during data collection \cite{wang2020multimodal}. 
%Medical imaging systems frequently encounter missing modalities due to equipment limitations or patient safety concerns \cite{WANGmedical}, and autonomous vehicles must operate reliably despite sensor failures \cite{park2025resilientsensorfusionadverse}. 
Traditional approaches address this through training-time strategies, either by training separate models for each modality combination or by employing modality dropout during training \cite{nezakati_MMP}. Yet both approaches are impractical for LVLMs, as retraining billion-parameter models for each missing modality scenario is computationally prohibitive (requiring $10^{20}$ FLOPs for 7B parameter models), while training-time modality dropout degrades full-modality performance and cannot adapt to deployment-specific missing patterns \cite{ma2021smil}.
\citet{ramazanova2025testtimeadaptationcombatingmissing} pose the missing modality as a domain shift problem and propose a method for TTA. The authors claim that despite advancements in the missing modality literature, a common drawback persists: all state-of-the-art approaches necessitate expensive retraining of the multimodal model. This poses a substantial challenge, particularly in applications with: (\textit{i}) extensive training data; (\textit{ii}) where the retraining
process is prohibitively expensive; and (\textit{iii}) when source data is not available, making the aforementioned approaches impractical. 

%We build upon this idea, but in the context of LVLMs. 
As shown with recent foundation models (e.g., GPT-style architectures and large multimodal systems \cite{videochatgpt2023,videollava2023}), retraining large language models (LLMs) and LVLMs is prohibitively expensive, requiring specialized hardware and weeks of distributed optimization. It is therefore impractical to retrain or fully fine-tune these models for every downstream task or deployment setting due to catastrophic forgetting, computational cost, and source data availability. This motivates the development of lightweight TTA methods for efficient customization without modifying the full parameter set. 
% Techniques such as parameter-efficient fine-tuning (PEFT), like low-rank adaptation (LORA), adapters, and prompt tuning methods, are commonly used to fine-tune a subset of model parameters on a smaller application dataset. 

%They allow for supervised adaptation to distribution shifts and task-specific requirements while preserving the pretrained backbone. Such approaches are essential for real-world deployment of LVLMs, where constraints on latency, memory, and energy demand scalable solutions that balance adaptability with computational complexity. 

\begin{figure}[!t]
    \centering
    \includegraphics[width=1.0\linewidth]{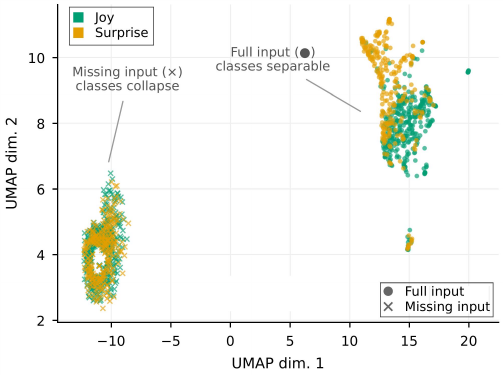}
    \caption{UMAP of the internal representations. The cross ($\times$) represents the missing modality inputs, and the circle ($\bullet$) represents the full modality inputs.}
    \label{fig:umap_layer26}
\end{figure}

%We formally 
We investigate the domain shift caused by missing modality inputs from 3 different viewpoints: (\textit{i}) UMAP visualization of the model's internal representations; (\textit{ii}) MMD computations between the full and missing modality inputs of the same sample; and (\textit{iii}) the degradation of performance. We perform this analysis on the MELD dataset with 50\% text-missing inputs using the \textit{Video-LLaVA-7B} model.
Figure~\ref{fig:umap_layer26} displays the representation geometry of the complete and text-missing inputs. The two colors correspond to the two classes. The $\bullet$ represents the full modality inputs and $\times$ represents the missing modality input for the same data samples. It reveals that the full and missing inputs are not only offset, but that missing inputs also blur the decision boundary. With full-modality inputs, the two classes form well-separated clusters, whereas under partial observation, this separation collapses entirely, with both classes overlapping into a single indistinguishable region

Quantitative analysis reveals a substantial internal distribution shift when modalities are missing, with an MMD of 0.8350 and cosine similarity of 0.6342 between complete and missing inputs for the same input samples. This indicates that the absence of modalities can significantly disturb the learned attention manifold, even though the semantic content remains unchanged. Lastly, the decline in performance shown in Tables~\ref{tab:meld_missing}-\ref{tab:bah_text_missing} indicates that LVLMs experience a severe performance drop (F1-score 0.6149 $\rightarrow $ 0.4785) in the 50\% text-missing case. Entropy-based TTA methods tend to drop towards near-random performance after adaptation. This observation is in line with the findings of \citet{hu2025testtimelearninglargelanguage}, which showed that entropy minimization objectives are ill-suited for the TTA of LLMs.

A second, less-addressed challenge in online TTA for LLMs and LVLMs is perpetual adaptation.
In practice, two failure modes emerge. First, once the student LoRA has converged, continued updates introduce gradient noise that perturbs the learned representation without improving performance ~\cite{wang2022cotta, niu2023stabletesttimeadaptationdynamic}.
Second, test streams in real deployments are non-stationary: speaker identities, recording conditions, and missing-modality rates shift over time.
Perpetual adaptation without drift awareness risks overwriting consolidated knowledge,
structurally analogous to catastrophic forgetting~\cite{Kirkpatrick_2017}.
Monitoring the raw distillation loss is an unreliable criterion for either stopping or restarting: the loss can plateau due to sample difficulty rather than genuine parameter convergence and can spike due to hard within-distribution samples rather than a true distributional shift.
This motivates two design choices: adaptation needs a supervision signal that fits autoregressive LVLMs, motivating latent alignment over entropy. A stopping signal grounded in the parameters themselves rather than the loss, motivating the Fisher information diagonal, whose shift from its value at convergence exposes genuine drift and lets adaptation pause and reactivate accordingly.

 The contributions of this paper are summarized as follows:
% \begin{enumerate}
% \textbf{(1)} A conditional latent representation alignment is introduced as a strong supervision signal for missing modality adaptation at test-time. 
% \textbf{(2)} TTSD: a parameter-efficient TTA method tailored to autoregressive LVLMs under missing modalities, overcoming the challenges of entropy-based objectives and retrieval methods. 
\textbf{(1)} TTSD: a parameter-efficient TTA method tailored to autoregressive LVLMs under missing modalities, built around conditional latent representation alignment as a supervision signal that overcomes the failure modes of entropy-based objectives and retrieval methods.
\textbf{(2)} FAR: a two-state stopping and drift-detection and reactivation mechanism for TTSD. It monitors the Fisher information diagonal to detect student convergence (\textsc{active} $\to$ \textsc{anchored}) and uses a Fisher Mismatch Index to detect test-stream distributional shift
(\textsc{anchored} $\to$ \textsc{active}).
\textbf{(3)} An extensive set of experiments on three challenging video-based ER datasets, namely MELD \cite{meld_ds}, DFEW \cite{dfew_ds}, and BAH \cite{gonzalez-26-bah}, with multiple missing ratios and across different modalities shows the effectiveness of our proposed TTSD-FAR method. 
% \end{enumerate}

%%%%%%%%%%%%%%%%%%%%%%%%%%%%%%%%%%%%%%%%%%%%%%%%%%%
\section{Related Works}
%%%%%%%%%%%%%%%%%%%%%%%%%%%%%%%%%%%%%%%%%%%%%%%%%%%
% MOVE to supplementary 
% \noindent {\textbf{Large Video–Language and Multimodal Autoregressive Models. }}
% Recent advances in multimodal foundation models extend vision–language pretraining to the video domain. Early generative frameworks such as VideoBERT~\cite{sun2019videobert} and UniVL~\cite{luo2020univl} align video and text via masked modeling, while modern instruction-tuned LVLMs like Video-ChatGPT~\cite{videochatgpt2023}, Video-Chat~\cite{li2023videochat}, and Video-LLaVA~\cite{videollava2023} integrate pretrained vision encoders and large language decoders for open-ended video understanding. These LVLMs achieve strong performance on multimodal tasks, but assume complete modality availability at inference, leading to distributional mismatch when modalities (e.g., transcripts, audio) are missing.

% \subsection{Multimodal Emotion Recognition}

% Emotion recognition traditionally employs multimodal fusion of visual, acoustic, and textual cues~\cite{poria2017review}, supported by datasets such as MELD~\cite{poria2019meld} and DFEW~\cite{jiang2021dfew}. Contrastive vision–language pretraining (e.g., CLIP~\cite{radford2021clip}) and multimodal LLM approaches~\cite{ge2024videoemotionopenvocabularyrecognition} further stimulate research, but still rely on full modality inputs. This limits applicability in real deployment scenarios where some modalities may be absent at test time.

%%%%%%%%%%%%%%%%%%%%%%%%%%%%%%%%%%%%%%%%%%%%%%%%%%%
\noindent {\textbf{Learning with Missing Modalities.}}
Multimodal learning with missing modalities has been extensively studied, with early works such as ModDrop~\cite{neverova2015moddrop}, SMIL~\cite{ma2021smil}, and masked modality projection (MMP)~\cite{nezakati_MMP} introducing train-time strategies to enhance robustness. \citet{maheshwari2024missing} propose a multi-modal teacher for masked modality learning, improving semantic segmentation under missing modalities. Some methods have explored modality-invariant architectures where single-branch models trained to be invariant to modality absence exhibit robustness during both training and testing~\cite{modalityinvariant2024, zhao2021missing,hemis2016}. \citet{guo2024multimodalprompt} use prompt learning in specialized cross-modal transformer architectures to regenerate the missing modality. More recently, \citet{Chen_Liu_Liu_Zhang_Li_Zhu_Liu_2026} proposed SMCIR, which detects sample-level modality missingness via an unsupervised entropy/mutual-information/similarity score (DMFD), then reconstructs missing modalities through context-guided multi-scale cross-modal attention (CMCG). Despite this progress, most methods rely on training-time strategies, either reconstructing missing modalities, simulating modality absence during training, or making custom architectural changes. These train‐time methods are not directly transferable to LVLMs, which are expensive to re-train and whose generative decoders lack clear self-supervised objectives that correlate with generation quality under modality absence.

%Move to supplementary
% Aslam \etal~\cite{aslam2023privileged,aslam2024multi} proposed privileged knowledge distillation (PKD) methods that leverage the learning using privileged information paradigm to transfer knowledge from multimodal teacher models, which have access to additional modalities during training, to unimodal student models that operate with a subset of modalities at test time. These methods assume that the privileged modality (e.g., physiological signal) is completely absent during deployment \cite{PKDOT}. In contrast, modalities like aural, visual, and textual are more commonly partially missing. Therefore, PKD methods do not directly generalize to adaptive test-time scenarios where generative cross-modal representations must be adjusted on the fly, motivating the need for methods like TTSD that adapt LVLMs at inference without retraining.

Multimodal systems often exhibit an imbalance in modality strength, with some modalities providing stronger task-relevant signals than others. When the stronger modality is absent, models experience a significant performance drop, highlighting the need for mechanisms that preserve knowledge from complete multimodal observations when operating even with weaker modalities.

\noindent {\textbf{Test-Time Adaptation.}}
TTA and test-time training have emerged to handle distribution shifts at inference time without access to target labels \cite{zhang2022memo,wang2022cotta}.
TENT proposes entropy minimization by adapting normalization statistics at test time~\cite{wang2021tent}. EATA~\cite{niu2022efficienttesttimemodeladaptation} improves entropy-based TTA by selectively updating model parameters using only reliable low-entropy samples and regularizing updates with a Fisher constraint to prevent forgetting. Many recent works analyze batch normalization for robustness under shift~\cite{nado2021evaluating}. A related line of continual-TTA work decides when to reset via drift detection rather than fixed schedules or entropy alone: periodic resets to pretrained weights curb long-horizon collapse~\cite{press2024rdumb}. Mishra ~\cite{rdumbpp2026} introduced RDumb++, a principled extension of RDumb that introduced two drift-detection mechanisms, i.e., entropy-based drift scoring and KL-divergence drift scoring, together with adaptive reset strategies. Alternating domain construction was proposed by~\cite{dsuta2024}. \citet{sun2020testtime} instead optimize auxiliary self-supervised losses during inference to improve adaptation. These methods are tailored to models that produce a single softmax distribution over a small, fixed label set, where output entropy is a direct, low-dimensional measure of prediction confidence. Self-supervised objectives and associated stopping signals do not transfer to LVLMs.
\citet{hu2025testtimelearninglargelanguage} introduced test-time learning for LLMs (TLM), adapting LLMs during inference using only unlabeled test data via input perplexity minimization rather than entropy minimization or supervised fine-tuning.
Adapting entire layers or normalization statistics in billion-parameter models further raises stability and efficiency concerns, motivating lightweight, parameter-efficient adaptation instead. Unlike EATA's per-step Fisher regularization, TTSD-FAR uses the Fisher diagonal as a control signal that suspends and resumes adaptation entirely.
\\
\noindent {\textbf{Retrieval-Augmented Generation.}}
RAG methods improve LLM and LVLM's performance during inference by incorporating external knowledge. Relevant information from a knowledge base is retrieved during inference \cite{asai2023selfraglearningretrievegenerate, jiang2024longragenhancingretrievalaugmentedgeneration, qian2025memoragboostinglongcontext}. RAG methods prove effective for tasks that require domain knowledge, but they rely heavily on the quality of retrieved information and incur additional computational latency. Precomputed vector databases or memory banks are also required for retrieval during inference \cite{abootorabi2025multimodalrag}. 

State-of-the-art methods to address missing modalities either require retraining with each modality combination or depend on discriminative learning objectives and specialized architectures to hallucinate missing modalities. The main issue faced by LVLMs for TTA is that there is no strong supervision signal available for adaptation. Entropy objectives do not provide meaningful supervision, and RAG schemes fail if the observed modality is too weak to retrieve correct class neighbors. This paper fills such a gap by proposing a novel approach for feature-level adaptation using dual LoRA adapters through self-distillation and FAR for convergence monitoring that allows stopping and resuming adaptation dynamically.

%%%%%%%%%%%%%%%%%%%%%%%%%%%%%%%
\section{Proposed Methodology}

Let $\mathcal{X} = \mathcal{X}^\text{v} \times \mathcal{X}^\text{t}$ denote the multimodal input space consisting of video and text modalities, where $\mathcal{X}^\text{v}$ represents video frames and $\mathcal{X}^\text{t}$ represents textual utterances. Let $\mathcal{Y}$ denote the output label space (e.g., emotion classes). We assume access to a pre-trained LVLM $f^{\theta} : \mathcal{X} \rightarrow \mathcal{Y},\ \theta \in \mathbb{R}^d,$ trained on large-scale multimodal corpora. The model consists of (i) modality-specific encoders $E^\text{v}: \mathcal{X}^\text{v} \rightarrow \mathbb{R}^{n \times d^\text{v}},\ E^\text{t}: \mathcal{X}^\text{t} \rightarrow \mathbb{R}^{m \times d^\text{t}},$ and (ii) a $\text{LLM}^\theta$ backbone $: \mathbb{R}^{(n+m) \times d} \rightarrow \mathcal{Y}$, which performs cross-modal fusion and autoregressive reasoning. We denote the frozen pre-trained backbone parameters by $\theta^0$ and the teacher and student LoRA weights by $\Delta\theta^\text{tea}=B^\text{tea}A^\text{tea}$ and $\Delta\theta^\text{stu}=B^\text{stu}A^\text{stu}$, respectively. We further denote the
vectorized student LoRA parameters as $\phi = \text{vec}(A^\text{stu}, B^\text{stu}) \in \mathbb{R}^p$, where $p$ is the total number of student LoRA parameters.

%%%%%%%%%%%%%%%%%%%%%%%%%%%%%%%%%%%%%%%%%%%%%%%%%%%
\begin{figure*}[tb]
  \centering
  \includegraphics[height=6cm]{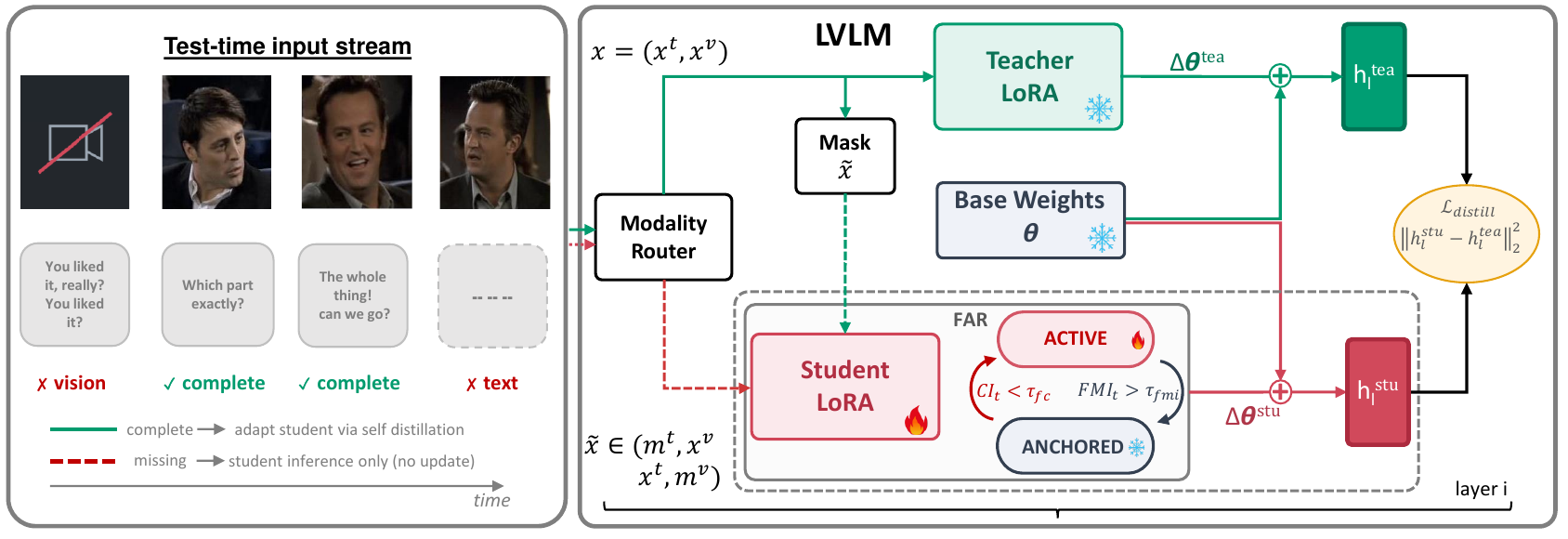}
  \caption{Illustration of the proposed TTSD-FAR method. The solid green arrow shows the flow of information in the case of complete modalities, and the dashed red arrow shows the flow of information in the missing modality case. The solid green arrow shows both modalities, and the single dashed green arrow denotes the input ($\tilde{x}$) with masked input. In the missing case, the model bypasses adaptation and only infers with the student LoRA. In the complete case, the model uses complete modality features from the teacher LoRA and masked modality features from the student LoRA, and distillation loss $\mathcal{L}_\text{distill}$ is calculated to match the student representation with the teacher representation.}
  \label{fig:ttsd_main}
\end{figure*}

\subsection{Problem Formulation}
%%%%%%%%%%%%%%%%%%%%%%%%%%%%%%%%%%%%%%%%%%%%%%%%%%%

\subsubsection{Latent Inference under Partial Observation.}

Let $x = (x^\text{v}, x^\text{t}) \sim P_{\text{full}}$ denote complete multimodal inputs
drawn from the training joint distribution. At test time, modalities may be missing. We
introduce binary modality masks $m^\text{v}, m^\text{t} \in \{0,1\},$ and define the
partially observed input as
\begin{equation}
    \tilde{x}
    =
    (\tilde{x}^\text{v}, \tilde{x}^\text{t})
    =
    (m^\text{v} \cdot x^\text{v},\; m^\text{t} \cdot x^\text{t}).
\end{equation}
The teacher's intermediate representation at layer $l$ is
\begin{equation}
    h^\text{tea}(x)
    =
    \mathrm{LLM}_{l}^{\theta^0 + \Delta\theta^\text{tea}}(x^\text{v}, x^\text{t})
    \in \mathbb{R}^{s \times d}.
\end{equation}
The missing-modality setting can be viewed as a \emph{partial observation problem}, where
the learner must infer a representation consistent with the latent multimodal structure
learned from complete data. We interpret $h^\text{tea}(x)$ as a latent variable lying on a multimodal representation manifold $\mathcal{Z}$. Under squared error loss, the Bayes-optimal estimator of the complete-modality representation given the observed input $\tilde{x}$ is the conditional expectation
\begin{equation}
    h^*(\tilde{x})
    =
    \mathbb{E}\!\left[ h^\text{tea}(x) \mid \tilde{x} \right].
\end{equation}

\subsubsection{TTSD as Conditional Latent Reconstruction.}

The student adapter parameterized by $\Delta\theta^\text{stu}$ produces
\begin{equation}
    h^\text{stu}(\tilde{x})
    =
    \mathrm{LLM}_{l}^{\theta^0 + \Delta\theta^\text{stu}}
    (\tilde{x}^\text{v}, \tilde{x}^\text{t}) \in \mathbb{R}^{s \times d}.
\end{equation}
TTSD minimizes the feature-level distillation objective
\begin{equation}
    \mathcal{L}_{\text{distill}}
    =
    \mathbb{E}_{x \sim P_{\text{full}}}
    \left[
    \left\| h^\text{stu}(\tilde{x}) - h^\text{tea}(x) \right\|_2^2
    \right],
\end{equation}
whose minimizer satisfies
${h^\text{stu}}^*(\tilde{x}) = \mathbb{E}\!\left[ h^\text{tea}(x) \mid \tilde{x} \right]$,
so that under a squared-error objective, the student is encouraged to approximate the
conditional expectation of the teacher representation given partial input.

However, perpetual adaptation of $\phi$ across an unbounded test stream introduces two
failure modes: (i) once $\Delta\theta^\text{stu}$ has converged, continued gradient updates add noise without improving the learned representation; and (ii) in non-stationary streams, parameter drift can overwrite previously consolidated knowledge. We therefore augment TTSD with Fisher-Anchored Restoration (FAR), a two-state mechanism that determines \emph{when} adaptation should proceed and \emph{when} it should be suspended, based solely on the geometry of the Fisher importance landscape.

%%%%%%%%%%%%%%%%%%%%%%%%%%%%%%%%%%%%%%%%%%%%%%%%%%%
\subsection{Architecture Overview}
%%%%%%%%%%%%%%%%%%%%%%%%%%%%%%%%%%%%%%%%%%%%%%%%%%%

Our TTSD-FAR framework operates on a single base video-language model $f^{\theta^0}$ with two
distinct LoRA adapters. Figure~\ref{fig:ttsd_main} illustrates the architecture. The
architecture enables efficient mode switching:
\begin{align}
    \text{Teacher mode:} \quad
        &\hat{y}^{\text{tea}} = f^{\theta^0 + \Delta\theta^\text{tea}}(x^\text{v}, x^\text{t}) \\
    \text{Student mode:} \quad
        &\hat{y}^{\text{stu}} = f^{\theta^0 + \Delta\theta^\text{stu}}(\tilde{x}^\text{v}, \tilde{x}^\text{t})
\end{align}

%%%%%%%%%%%%%%%%%%%%%%%%%%%%%%%%%%%%%%%%%%%%%%%%%%%
\subsection{TTA: Student LoRA}
%%%%%%%%%%%%%%%%%%%%%%%%%%%%%%%%%%%%%%%%%%%%%%%%%%%
At test time, when encountering a sample ${x}$ with complete modalities, we perform online adaptation of the student LoRA through feature-level self-distillation from the teacher. By directly matching intermediate representations, the student adapter is encouraged to capture the internal geometry of the teacher representation space rather than imitating its final decisions. This preserves the diversity and richness of the teacher features and mitigates mode collapse, which is particularly critical under missing-modality scenarios.
% TTSD is inherently modality-agnostic: the supervision signal depends solely on the
% alignment of latent representations, independent of which input modality is absent.

Given a layer $l\in\mathcal{L}$ of the LLM backbone, the hidden representations are:
\begin{align}
    h^\text{tea}_l &= \text{LLM}_l^{\theta^0 + \Delta\theta^\text{tea}}
        (E^\text{v}(x^\text{v}), E^\text{t}(x^\text{t})) \in \mathbb{R}^{s \times d} \\
    h^\text{stu}_l &= \text{LLM}_l^{\theta^0 + \Delta\theta^\text{stu}}
        (E^\text{v}(\tilde{x}^\text{v}), E^\text{t}(\tilde{x}^\text{t})) \in \mathbb{R}^{s \times d}
\end{align}
where $s$ is the sequence length (the sum of video tokens and text tokens) and $d$ is the hidden dimension. We minimize the mean squared error between student and teacher representations as:
\begin{equation}
    \mathcal{L}_{\text{distill}}
    =
    \sum_{l \in \mathcal{L}} \alpha_l \cdot
    \| h^\text{stu}_l - \text{sg}(h^\text{tea}_l) \|_2^2
    \label{eq:distill_loss}
\end{equation}
where $\text{sg}(\cdot)$ denotes the stop-gradient operation to prevent backpropagation into the frozen teacher, and $\alpha_l$ are layer-specific weights. 

Teacher features are extracted once per sample. To ensure complete isolation, features are detached from the teacher network because no gradient calculation is required in the teacher LoRA. Teacher and student forward passes do not share computational graph nodes, preventing gradient contamination. Without isolation, shared normalization statistics or attention caches can cause interference.

\begin{tcolorbox}[colback=blue!5, colframe=blue!35!black, arc=3pt, left=1mm, right=1mm]

\noindent\textbf{Why Self-Distillation for Missing Modalities?}

\noindent The teacher, trained on complete modalities, has learned rich cross-modal features that fuse information from both vision and text. When a modality is missing at test time, the student can be viewed as solving an inverse problem:
\begin{equation}
\Delta\theta^{\mathrm{stu}}
=
\arg\min_{\Delta\theta}
\mathbb{E}_{\tilde{x}}
\left\|
\mathbf{h}_{\mathrm{stu}}(\tilde{x})
-
\mathbf{h}_{\mathrm{tea}}(x)
\right\|_2^2.
\label{eq:main_opt}
\end{equation}

\end{tcolorbox}

% \begin{tcolorbox}[colback=blue!5, colframe=blue!35!black, arc=3pt]
% \noindent\textbf{Why Self-Distillation Works for Missing Modalities?}

% \noindent The teacher, trained on complete modalities, has learned rich cross-modal features
% that fuse information from both vision and text. When a modality is missing at test time, the
% student can be viewed as solving an inverse problem:
% \begin{equation}
%     \Delta\theta^\text{stu}
%     =
%     \arg\min_{\Delta\theta}
%     \mathbb{E}_{\tilde{x}}
%     \left\|
%     \text{LLM}_l^{\theta^0+\Delta\theta}(\tilde{x})
%     -
%     \text{LLM}_l^{\theta^0 + \Delta\theta^\text{tea}}(x)
%     \right\|_2^2
%     \label{eq:main_opt}
% \end{equation}
% \end{tcolorbox}

%%%%%%%%%%%%%%%%%%%%%%%%%%%%%%%%%%%%%%%%%%%%%%%%%%%
\subsection{Fisher-Anchored Restoration (FAR)}
\label{sec:far}
%%%%%%%%%%%%%%%%%%%%%%%%%%%%%%%%%%%%%%%%%%%%%%%%%%%

\noindent Continuous adaptation of $\phi$ is not harmless: once the student has found a
solution to Eq.~(\ref{eq:main_opt}), additional gradient steps on the same distribution
degrade performance through noise accumulation, while distribution shifts in the test stream
can silently overwrite consolidated knowledge. Monitoring $\mathcal{L}_{\text{distill}}$ directly
is unreliable as a stopping signal since the loss can plateau not because the student has
converged, but because the current batch of samples is uniformly hard. The correct stopping
signal is the stability of the \emph{parameter importance landscape}. Specifically, whether
the curvature of $\mathcal{L}_{\text{distill}}$ with respect to $\phi$ has stabilised.
Symmetrically, drift should not be detected by raw loss spikes, which can be triggered by
hard samples in a stationary stream. Instead, drift is detected by monitoring whether
incoming gradients engage parameter dimensions that were \emph{unimportant} during
consolidation. FAR formalizes this intuition through two complementary Fisher-based statistics and a
two-state mechanism:
$\text{\sc active}$ (the student LoRA is being updated) and
$\text{\sc anchored}$ (the student LoRA is frozen).

\subsubsection{Online Fisher Diagonal.}

After each adaptation step, the diagonal of the empirical Fisher information matrix over
$\phi$ is updated via an exponential moving average:
\begin{equation}
    F_t
    =
    \beta_F \cdot F_{t-1}
    +
    (1 - \beta_F) \cdot
    \left( \nabla_{\phi}\,\mathcal{L}_{\text{distill}} \right)^{\odot 2}
    \label{eq:fisher_ema}
\end{equation}
where $(\cdot)^{\odot 2}$ denotes element-wise squaring, $\beta_F \in (0,1)$ is the
decay parameter controlling the weight placed on historical estimates. The element $F_{t,j}$ approximates the
expected squared gradient with respect to the $j$-th parameter in $\phi$, providing a
running estimate of each parameter's contribution to the distillation objective.
$F_t$ is updated only during the $\text{\sc active}$ state.

\subsubsection{Consolidation Index and the \textsc{active} $\to$ \textsc{anchored} Transition.}

The consolidation index (CI) measures the normalized $\ell_1$ rate of change of the Fisher diagonal between consecutive adaptation steps:
\begin{equation}
    \mathrm{CI}_t
    =
    \frac{\| F_t - F_{t-1} \|_1}{\| F_t \|_1 + \varepsilon}
    \label{eq:ci}
\end{equation}
where $\varepsilon > 0$ is a small numerical stability constant. When
$\mathrm{CI}_t$ is large, the importance landscape is still evolving, i.e., the student has not yet settled on a stable curvature configuration. As the student converges,
$\mathrm{CI}_t \to 0$, indicating that the Fisher diagonal has stopped shifting.
The $\ell_1$ norm is chosen because $F_t$ is a non-negative vector, and $\ell_1$
gives the total mass shifted in the importance landscape, which has a direct
interpretation as the number of parameters whose importance is still changing. 
The transition from $\text{\sc active}$ $\rightarrow$ $\text{\sc anchored}$ is triggered when
$\mathrm{CI}_t$ remains below a threshold $\tau_\text{fc}$ for $K$ consecutive adaptation steps $\mathrm{CI}_t < \tau_\text{fc}$.
% \begin{equation}
%     \text{\sc active}
%     \;\rightarrow\;
%     \text{\sc anchored}
%     \quad \text{when} \quad
%     \mathrm{CI}_t < \tau_\text{fc}
%     \;\; \text{for} K \text{ cons. steps}
%     \label{eq:consolidation}
% \end{equation}
% At the transition, the Fisher anchor is stored:
% \begin{equation}
    % F_\text{anchor}
    % =F_t
    % \label{eq:anchor}
% \end{equation}
The student LoRA parameters $\phi$ are
then frozen; no weight updates are applied while in the $\text{\sc anchored}$ state, and the current $F_t$ is stored as an anchor, $F_\text{anchor}=F_t$ .

\subsubsection{Fisher Mismatch Index and the \textsc{anchored} $\to$ \textsc{active} Transition.}

While $\text{\sc anchored}$, for each incoming complete-modality sample, the gradient of
$\mathcal{L}_{\text{distill}}$ is computed at the frozen anchor parameters without applying
a weight update:
\begin{equation}
    g_t
    =
    \nabla_{\phi}\,\mathcal{L}_{\text{distill}}
    \Big|_{\phi = \phi_\text{anchor}}
    \label{eq:fmi_grad}
\end{equation}
The Fisher mismatch index (FMI) for sample $t$ is then
\begin{equation}
    \mathrm{FMI}_t
    =
    \frac{1}{p}
    \sum_{j=1}^{p}
    \frac{g_{t,j}^{2}}{F_{\text{anchor},j} + \varepsilon}
    \label{eq:fmi}
\end{equation}
where $p = |\phi|$ is the total number of student LoRA parameters. The ratio
$g_{t,j}^2 / F_{\text{anchor},j}$ is large when the current sample induces a strong
gradient in parameter $j$, but that parameter had low importance during consolidation.
This points to large gradients in directions the consolidated student found unimportant. 
The transition from $\text{\sc anchored}$ back to $\text{\sc active}$ 
is triggered when ${\mathrm{FMI}}_{\,t} > \tau_\text{fmi}$ exceeds a threshold.
% \begin{equation}
%     \text{\sc anchored}
%     \;\longrightarrow\;
%     \text{\sc active}
%     \quad \text{when} \quad
%     {\mathrm{FMI}}_{\,t} > \tau_\text{fmi}.
%     \label{eq:drift}
% \end{equation}

% \subsubsection{FAR Hyperparameters.}

% \noindent Table~\ref{tab:far_hparams} summarises the FAR hyperparameters, their roles, and
% their suggested priors. All thresholds are dimensionless and interpretable: $\tau_\text{fc}$
% is a percentage change in the Fisher diagonal, $K$ is a step count, $M$ is a window length,
% and $\tau_\text{fmi}$ is a ratio of current to consolidated parameter importance.

% \begin{table}[h]
% \centering
% \caption{FAR hyperparameters.}
% \label{tab:far_hparams}
% \small
% \begin{tabular}{llcc}
% \toprule
% \textbf{Symbol} & \textbf{Role} & \textbf{Prior} \\
% \midrule
% $\beta_F$         & EMA decay for Fisher diagonal              & 0.99 \\
% $\tau_\text{fc}$  & CI threshold for consolidation             & 0.02 \\
% $K$               & Consecutive low-CI steps to trigger anchor & 5    \\
% $M$               & FMI sliding window size                    & 20   \\
% $\tau_\text{fmi}$ & FMI threshold for drift detection          & 2.0  \\
% $\beta_\text{reset}$ & Fraction of $F_\text{anchor}$ discarded on restoration & 0.3 \\
% \bottomrule
% \end{tabular}
% \end{table}

%%%%%%%%%%%%%%%%%%%%%%%%%%%%%%%%%%%%%%%%%%%%%%%%%%
\section{Results and Discussion}
\begin{table*}[t]
\centering
\footnotesize
\setlength{\tabcolsep}{4.6pt}
\caption{Comparison of methods under progressively missing textual and visual modalities. We report F1 on the MELD dataset as the proportion of unavailable input increases. N/A: Perplexity Gen. is not applicable to the vision-missing case.}
\label{tab:meld_missing}
\begin{tabular}{lc ccccc ccccc}
\toprule
& & \multicolumn{5}{c}{\textbf{Text Missing}} & \multicolumn{5}{c}{\textbf{Vision Missing}} \\
\cmidrule(lr){3-7} \cmidrule(lr){8-12}
\textbf{Method} & \textbf{Full} & \textbf{10\%} & \textbf{20\%} & \textbf{30\%} & \textbf{40\%} & \textbf{50\%} & \textbf{10\%} & \textbf{20\%} & \textbf{30\%} & \textbf{40\%} & \textbf{50\%} \\
\midrule
No Adaptation & 0.6149 & 0.5926 & 0.5706 & 0.5390 & 0.5190 & 0.4785 & 0.6122 & 0.6025 & 0.5934 & 0.5825 & 0.5775 \\
TENT{\fontsize{6}{12} \selectfont (ICLR'21)} & -- & 0.1650 & 0.1680 & 0.1765 & 0.1855 & 0.1655 & 0.1658 & 0.1725 & 0.1655 & 0.1838 & 0.1826 \\
EATA{\fontsize{6}{12} \selectfont (ICML'22)} & -- & 0.2854 & 0.2690 & 0.2770 & 0.2736 & 0.2482 & 0.2950 & 0.2750 & 0.2745 & 0.2530 & 0.2382 \\
RAG{\fontsize{6}{12} \selectfont (ICCV'25)} & -- & 0.5640 & 0.5248 & 0.5120 & 0.4845 & 0.4660 & 0.5925 & 0.5830 & 0.5634 & 0.5567 & 0.5534 \\
Perplexity Gen.   & -- & 0.5680 & 0.5460 & 0.5255 & 0.4938 & 0.4722 & N/A & N/A & N/A & N/A & N/A \\
\textbf{TTSD-FAR (Ours)} & -- & \textbf{0.6115} & \textbf{0.5890} & \textbf{0.5625} & \textbf{0.5454} & \textbf{0.5124} & \textbf{0.6130} & \textbf{0.6075} & \textbf{0.6055} & \textbf{0.5970} & \textbf{0.5915} \\
\bottomrule
\end{tabular}
\end{table*}

%%%%%%%%%%%%%%%%%%%%%%%%%%%%%%%%%%%%%%%%%%%%%%%%%%%
\subsection{Experimental Methodology}

\noindent {\textbf{Datasets. }}
We evaluate our approach on three widely used multimodal and video-based emotion recognition benchmarks: Multimodal Emotion Lines Dataset \textbf{(MELD)} \cite{meld_ds}, Dynamic Facial Expression in-the-Wild \textbf{(DFEW)} \cite{dfew_ds}, and Behavioral Ambivalence/Hesitancy \textbf{(BAH)} \cite{gonzalez-26-bah}, each presenting distinct challenges in terms of modality diversity, temporal dynamics, and subtle affective cues. Details of the datasets are provided in the supplementary material. We exclusively validate on emotion recognition, which is one of the few multimodal applications where text transcripts serve as a crucial, non-redundant input modality. This makes it an ideal testbed for missing-modality robustness, since dropping the text modality induces genuine information loss.

\noindent {\textbf{Evaluation Protocol.}}
This paper makes no assumptions about the training phase. TTSD-FAR only requires a model that works with multiple modalities at test time. To the best of our knowledge, this is the first TTA method that effectively adapts to missing modalities in LVLMs. TTSD-FAR updates the model only when it encounters modality-complete samples but generates predictions for all samples, regardless of the modalities they contain. Missing modality scenarios are simulated by randomly masking modalities at inference time, with missing ratios ranging from 0\% (complete) to 50\%. All baselines use the same frozen LVLM backbone for fair comparison. Detailed implementation settings for each dataset are provided in the supplementary material.

%%%

% \subsection{Results on the MELD Dataset}
\label{subsec:textmissmeld}
\subsection{Results under Missing Modalities}

Tables~\ref{tab:meld_missing}--~\ref{tab:bah_text_missing} summarize performance across MELD, DFEW, and BAH under progressively increasing missing-modality ratios. Across all three datasets, entropy-based TTA methods (TENT and EATA) consistently underperform, confirming that confidence-based objectives provide unreliable supervision for autoregressive LVLMs under severe modality shift. This observation is consistent with the findings of \citet{hu2025testtimelearninglargelanguage}, who showed that entropy minimization does not provide reliable optimization signals for autoregressive language models at test time. RAG and perplexity-based generation remain competitive when the observed modality is sufficiently informative, but their performance degrades as the dominant modality becomes unavailable.

On MELD (Table~\ref{tab:meld_missing}), removing text causes substantially greater degradation than removing vision ($0.6149 \rightarrow 0.4785$ versus $0.6149 \rightarrow 0.5775$), indicating that textual information provides the dominant supervision signal for emotion recognition in conversational settings. Consequently, retrieval-based methods struggle once textual information is unavailable, while entropy-based methods rapidly collapse toward near-random performance. TTSD-FAR instead leverages complete-modality teacher representations to supervise a student operating on synthetically masked inputs, allowing it to remain close to the full-modality baseline across both missing-text and missing-vision settings. A similar trend is observed on DFEW (Table~\ref{tab:dfew_vision_missing}), where removing visual information induces a substantial performance drop because facial dynamics constitute the dominant modality. Entropy-based adaptation again fails, while RAG deteriorates as weaker visual representations reduce retrieval quality. TTSD-FAR consistently achieves the strongest performance across all missing ratios, demonstrating that feature-level self-distillation provides a substantially more reliable adaptation signal than confidence- or retrieval-based approaches. Results for the weaker text-missing scenario are included in the supplementary material.
Results on BAH (Table~\ref{tab:bah_text_missing}) further demonstrate the generality of the proposed framework. Although entropy-based methods perform relatively better because BAH is a binary classification task, they remain consistently inferior to TTSD-FAR. RAG and perplexity-based generation improve upon the zero-shot baseline at lower missing ratios but gradually deteriorate as textual information becomes increasingly unavailable. Additional implementation details and supplementary experiments are provided in the supplementary material.

\begin{table}[t]
\centering
\footnotesize
\setlength{\tabcolsep}{1.5pt}
\caption{Comparison of methods under progressively missing visual modality. We report F1 on the DFEW dataset as the proportion of unavailable visual input increases.}
\label{tab:dfew_vision_missing}
\begin{tabular}{lcccccc}
\toprule
\textbf{Method} & \textbf{Full} & \textbf{10\%} & \textbf{20\%} & \textbf{30\%} & \textbf{40\%} & \textbf{50\%} \\
\midrule
No Adaptation & \textbf{0.5549} & 0.5300 & 0.5078 & 0.4984 & 0.4837 & 0.4434 \\
\midrule
TENT & -- & 0.1548 & 0.1553 & 0.1665 & 0.1658 & 0.1640 \\
EATA & -- & 0.2534 & 0.2778 & 0.2735 & 0.2845 & 0.2885 \\
RAG & -- & 0.5340 & 0.5048 & 0.5020 & 0.4800 & 0.4468 \\
\textbf{TTSD-FAR (Ours)} & -- & \textbf{0.5515} & \textbf{0.5300} & \textbf{0.5195} & \textbf{0.5015} & \textbf{0.4775} \\
\bottomrule
\end{tabular}
\end{table}

%%%%%%%%%%
% \subsection{Results on the BAH Dataset}
% \noindent\textbf{BAH: Text Missing at Test Time.}
% This section presents the results for the text missing from the BAH dataset. Similar to MELD and DFEW, the TENT and EATA methods negatively impact model performance. However, because ambivalence recognition is a binary classification task, performance achieved by entropy minimization methods is considerably higher than in Tables~\ref{tab:meld_missing} and~\ref{tab:dfew_vision_missing}, which are 7-class classification problems. The RAG and perplexity-based text generation methods are able to improve upon the zero-shot performance in some cases. In all cases, the TTSD-FAR is able to outperform the existing TTA methods and significantly improve model performance under missing modality conditions. The ambivalence/hesitancy recognition is a specialized task that requires carefully tailored prompts. Implementation details of the ambivalence/hesitancy task are presented in the supplementary material. 

\begin{table}[t]
\centering
\footnotesize
\setlength{\tabcolsep}{1.5pt}
\caption{Comparison of methods under progressively missing textual modality conditions for BAH dataset.}
\label{tab:bah_text_missing}
\begin{tabular}{lcccccc}
\toprule
\textbf{Method} & \textbf{Full} & \textbf{10\%} & \textbf{20\%} & \textbf{30\%} & \textbf{40\%} & \textbf{50\%}\\
\midrule
Zero-Shot & \textbf{0.7142} & 0.6835 & 0.6742 & 0.6526 & 0.6215 & 0.6023\\
\midrule
TENT & -- & 0.4923 & 0.4834 & 0.4710 & 0.4572 & 0.4550\\
EATA & -- & 0.5630 & 0.5467 & 0.5360 & 0.5310 & 0.5250\\
RAG & -- & 0.6835 & 0.6730 & 0.6535 & 0.6215 & 0.5950\\
Perplexity Gen. & -- & 0.6870 & 0.6745 & 0.6535 & 0.6310 & 0.6030\\
\textbf{TTSD-FAR (Ours)} & -- & \textbf{0.7050} & \textbf{0.6925} &
\textbf{0.6695} & \textbf{0.6405} & \textbf{0.6250}\\
\bottomrule
\end{tabular}
\end{table}

%%%%%%%%%
\subsection{Computational Complexity Analysis }
Experiments show that TTSD-FAR performs well when adapting to missing modality conditions at test time. However, this performance gain comes at a computational cost. We distinguish the two cases and explain the computational overhead for both cases.
i) \textbf{Full modality case:} TTSD-FAR requires two forward passes, one forward pass for obtaining the teacher features and one for obtaining the student features. ii) \textbf{Missing modality case:} For the missing modality case, there is no computational overhead. The model uses the adapted student LoRA for prediction. We update only 44M parameters out of the 7B parameters of the \textit{Video-LLaVA 7B} model. This results in only $\approx$ 0.629\% of the parameter updates. It is also important to note that the total wall time for the adaptation scenario depends on the missing ratio. For the 50\% missing case, 50\% of the samples will pose no additional overhead; the remaining 50\% of the input samples require two forward passes and backpropagation to the student LoRA. Consequently, the total wall time of the complete MELD test set without adaptation takes $\approx$ 1.5 hours, and the time taken with adaptation is $\approx$ 2.1 hours.

%%%%%%%%%%%%
\subsection{Mechanistic Analysis of FAR}

To isolate the contribution of FAR, we compare TTSD with and without the restoration mechanism across MELD and BAH under increasing missing rates.
\begin{figure}[!t]
  \centering
  \includegraphics[height=4.1cm]{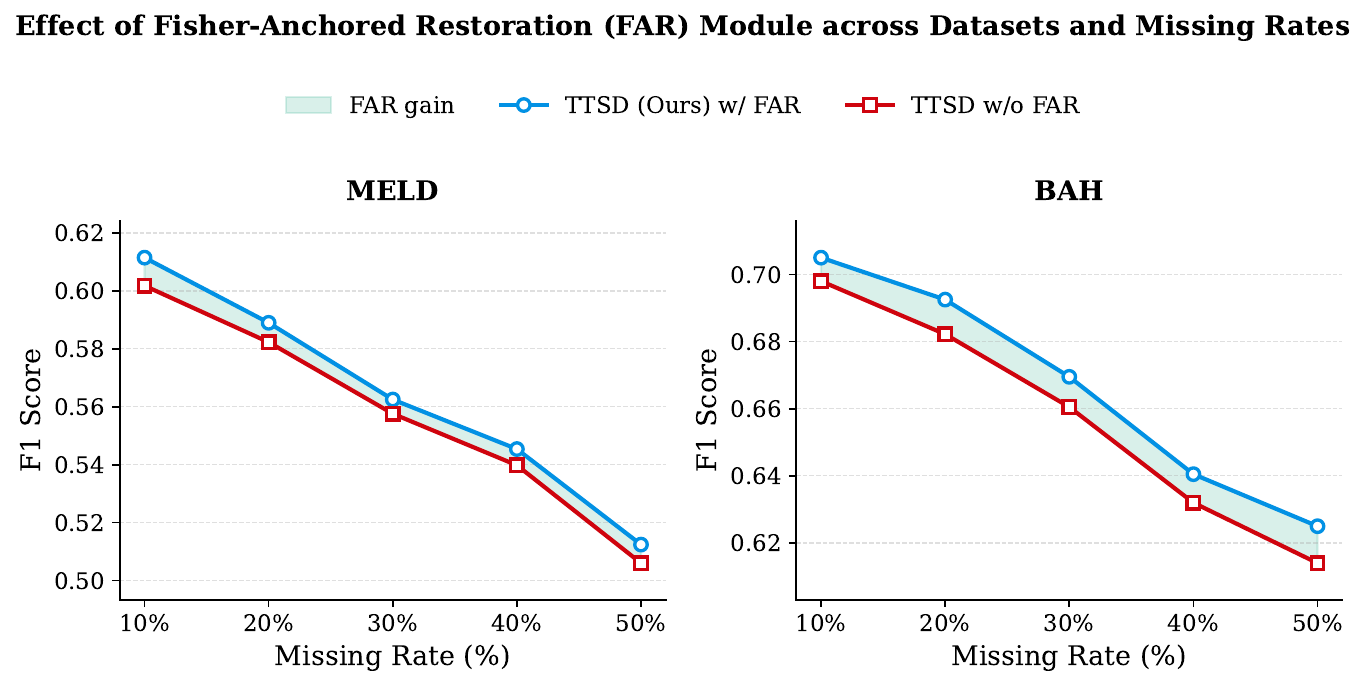}
  \caption{Effectiveness of the FAR module on the distillation results across different missing rates.}
  \label{fig:far_ablation}
\end{figure}

As shown in Figure~\ref{fig:far_ablation}, the unbounded variant (TTSD w/o FAR), which continues to adapt the student LoRA on every incoming sample, consistently underperforms the FAR-governed version at every missing rate. The consistent FAR gain across both the 7-class (MELD) and binary (BAH) settings indicates that this stopping/restarting behavior is not task-specific; rather, it reflects a general property of self-distillation under non-stationary missing-modality streams.

\begin{figure}[!t]
  \centering
  \includegraphics[height=2.85cm]{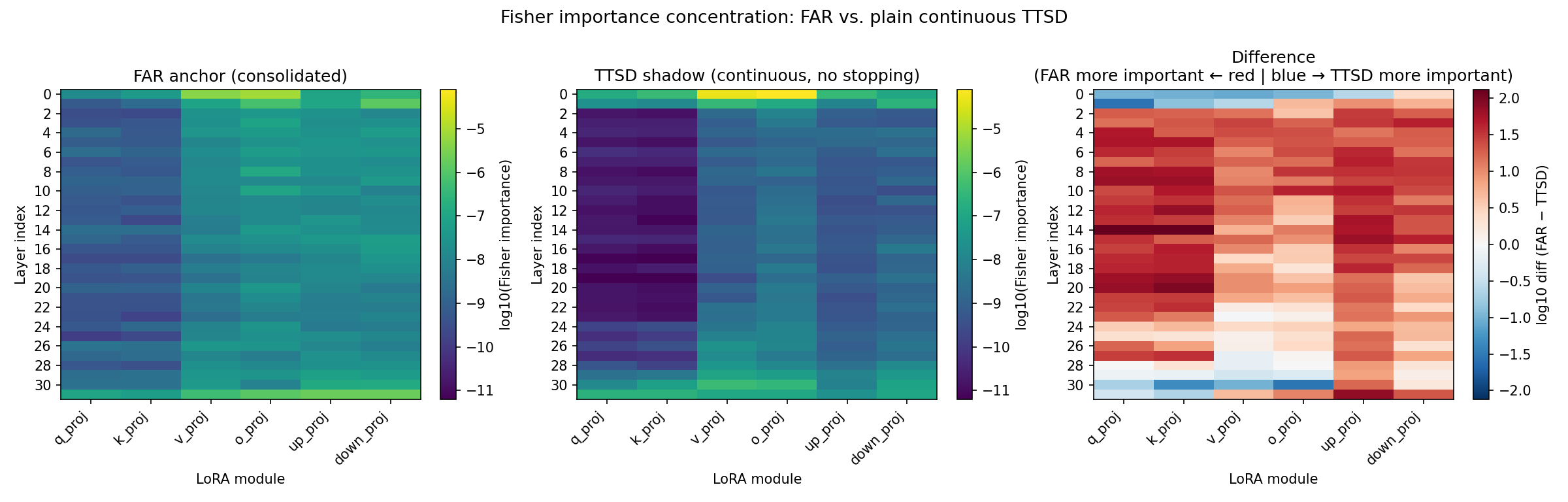}
  \caption{Fisher importance heatmaps (log$_{10}$ scale): TTSD-FAR (left), TTSD (middle), and difference map (right), confirming systematically higher importance under FAR.}
  \label{fig:far_heatmap}
\end{figure}

The three heatmaps in Figure~\ref{fig:far_heatmap} show how Fisher parameter importance is distributed across LoRA modules under two adaptation regimes. The left panel shows $F_\text{anchor}$, the Fisher diagonal stored at the moment FAR declared convergence, where importance is sharply concentrated in the early layers of the query and up-projection modules, with the remainder of the landscape several orders of magnitude lower; this sparsity indicates that the student identified a precise, low-dimensional parameter subspace sufficient to solve the missing-modality distillation task. The middle panel shows the equivalent Fisher diagonal accumulated by continuous TTSD without any stopping criterion. The importance landscape is substantially more diffuse, and the gradient mass is spread uniformly across all layers and modules, reflecting that continued adaptation beyond convergence dilutes the signal that was meaningful at true convergence.
The right panel shows the log-scale difference, which is almost entirely red, confirming that FAR anchors at a moment of higher and more concentrated importance across most of the parameter space.

\subsection{Discussion}
Our experiments reveal two core challenges for adapting autoregressive LVLMs under partial modality observation: recovering missing semantic content and sustaining effective adaptation over long deployment.
Retrieval-based adaptation assumes the observed modality provides a query informative enough to retrieve semantically relevant references. This assumption breaks down when the missing modality carries the dominant information, in which case the retrieved examples supply inconsistent or misleading supervision.
Entropy minimization fails in a complementary way. It implicitly assumes confident predictions are correct predictions, but under substantial modality shift, autoregressive LVLMs often generate overconfident, semantically incorrect outputs. Minimizing entropy in this regime reinforces these errors, producing optimization drift rather than genuine adaptation, consistent with the findings of \citet{hu2025testtimelearninglargelanguage}.

TTSD-FAR addresses both failure modes by treating missing-modality adaptation as a conditional representation alignment problem rather than a confidence or retrieval optimization problem. The student is guided toward the frozen reference model's complete-modality latent representation through feature-level supervision, recovering semantic information that the observed modality alone cannot provide and yielding substantially more robust adaptation under severe degradation.
A separate issue is that continued optimization past convergence causes parameter diffusion, eroding useful representations. FAR's adaptive consolidation addresses this by tracking optimization stationarity via Fisher geometry to detect convergence, then reactivating adaptation only when deviations from the consolidated geometry signal genuine distributional drift. This preserves learned representations while retaining the capacity to adapt when the input distribution actually shifts.

\section{Conclusion}
TTSD-FAR is a parameter-efficient framework for adapting large video-language models to missing modalities during inference. Missing modalities induce substantial shifts in representation geometry within autoregressive LVLMs, a regime in which entropy-based test-time adaptation and retrieval-augmented generation both fail to provide reliable supervision. TTSD-FAR instead couples a frozen teacher, trained on complete modalities, with an adaptive student adapter that operates on masked inputs, aligning student and teacher representations via feature-level self-distillation without modifying the base model's parameters. Fisher-Anchored Restoration governs this process through a cycle of consolidation and reactivation: adaptation is suspended once the student's parameter importance landscape stabilizes, and resumed only when the Fisher geometry signals genuine distributional drift, preventing the parameter diffusion that unbounded adaptation otherwise induces. Extensive experiments across three emotion recognition benchmarks show that TTSD-FAR consistently improves performance under severe missing-modality conditions while preserving accuracy competitive with the complete-modality setting.
%In the future, this fundamental method can be extended to noisy modalities as well. The masking module can be replaced with a pretrained modality translator that translates the source modality to the target modality. Another highly impactful direction would be to work on the adapter merging after TTA. Although the proposed method only incurs a very light overhead, adapter merging can eliminate the overhead. 

\noindent\textbf{Supplementary materials. } 
Includes proof sketches, dataset description and implementation details, algorithm for the proposed TTSD-FAR method, and extended ablation studies.

\appendix

\section{Appendix}
\label{sec:appendix}
%%%%%%%%%%%%%%%%%%%%%%%%%%%%%%%%%%%%%%%%%%%%%%%%%%%
% --- Manual Table of Contents ---
\newcommand{\tocsec}[2]{\noindent\textbf{#1}\dotfill\pageref{#2}\par\vspace{2pt}}
\newcommand{\tocsub}[2]{\noindent\hspace*{1em}#1\dotfill\pageref{#2}\par\vspace{1pt}}
\newcommand{\tocsubsub}[2]{\noindent\hspace*{2em}\textit{#1}\dotfill\pageref{#2}\par}

\begin{flushleft}

\tocsec{Algorithm for TTSD-FAR}{alg:far}

\tocsec{Theoretical Properties of FAR}{sec:far-theory}
% \tocsubsub{Assumptions}{sec:far-assumptions}
\tocsubsub{Prop. 1: Fisher Stability as a Stationarity Signal}{prop:fisher-stability}
\tocsubsub{Proposition 2: No-Regret Reactivation}{prop:no-regret}
\tocsubsub{Proposition 3: FMI Hard-Sample Robustness}{prop:fmi-robustness}

\vspace{4pt}
\tocsec{Additional Results}{sec:additional-results}
\tocsub{Distilling from Base Teacher (no teacher LoRA)}{subsec:distill-notl}
\tocsub{Weaker Modality Missing Results}{sup-sec:dataset-imp}
% \tocsubsub{DFEW -- Text Missing at Test-Time -- Weaker Modality Missing}{sec:dfew-text-missing}
% \tocsubsub{BAH -- Vision Missing at Test-Time -- Weaker Modality Missing}{sec:bah-vision-missing}

\vspace{4pt}
\tocsec{Additional Ablations}{sec:additional-ablations}
\tocsub{LoRA Rank Sensitivity}{sec:lora-rank-sensitivity}
\tocsub{Effectivity of TTSD on Small Video Language Model}{sec:svlm}
\tocsub{FAR Hyperparameters Sensitivity}{sec:far-hyperparam-sensitivity}
\tocsubsub{Effect of the FMI Threshold}{fig:ttsd_fmiplot}
\tocsubsub{FAR State Dynamics over the Test Stream}{sec:far-state-dynamics}
\tocsub{Loss-Based Stopping vs.\ FAR}{tab:loss-stop-ablation}

\vspace{4pt}
\tocsec{Datasets and Implementation Details}{sec:datasets-impl}
\tocsub{Datasets}{sec:datasets}
\tocsub{Implementation Details}{sec:impl-details}
% \tocsubsub{Missing Modality Simulation}{fig:missing_img}
\tocsubsub{FAR Parameters}{tab:far-params-brief}

\end{flushleft}
% --- End Manual Table of Contents ---

\section{Algorithm for TTSD-FAR}
Algorithm \ref{alg:far} summarizes the adaptation loop for the LVLM using the proposed TTSD-FAR methodology.
\begin{algorithm}[h]
\caption{TTSD-FAR}
\label{alg:far}
\begin{algorithmic}[1]
\REQUIRE $\Delta\theta^{\text{tea}}$, $\phi_0$, $\eta$, $\beta_F$, $\tau_{\text{fc}}$, $K$, $\tau_{\text{fmi}}$, $M$, $\varepsilon$
\STATE $\phi \leftarrow \phi_0$;\ $F, F_{\text{prev}} \leftarrow \mathbf{0}_p$;\ $\text{state} \leftarrow \textsc{Active}$;\ $c \leftarrow 0$;\ $\mathcal{B} \leftarrow \emptyset$
\FOR{each incoming sample $\tilde{x}$}
    \IF{$\text{state} = \textsc{Active}$}
        \STATE $h^{\text{tea}}_l \leftarrow \mathrm{sg}(\text{LLM}_l^{\theta^0+\Delta\theta^{\text{tea}}}(x))$;\ $h^{\text{stu}}_l \leftarrow \text{LLM}_l^{\theta^0+\Delta\theta^{\text{stu}}}(\tilde{x})$
        \STATE $g \leftarrow \nabla_{\phi} \sum_{l} \alpha_l \|h^{\text{stu}}_l - h^{\text{tea}}_l\|_2^2$;\ $\phi \leftarrow \phi - \eta g$
        \STATE $F \leftarrow \beta_F F_{\text{prev}} + (1-\beta_F) g^{\odot 2}$
        \STATE $\mathrm{CI} \leftarrow \|F - F_{\text{prev}}\|_1 / (\|F\|_1 + \varepsilon)$;\ $F_{\text{prev}} \leftarrow F$
        \STATE $c \leftarrow c+1$ if $\mathrm{CI} < \tau_{\text{fc}}$, else $c \leftarrow 0$
        \IF{$c \geq K$}
            \STATE $\phi_{\text{anchor}}, F_{\text{anchor}} \leftarrow \phi, F$;\ $\text{state} \leftarrow \textsc{Anchored}$;\ $\mathcal{B} \leftarrow \emptyset$
        \ENDIF
    \ELSE
        \STATE Infer with frozen $\phi_{\text{anchor}}$ \COMMENT{no update}
        \IF{$\tilde{x}$ is a complete-modality sample}
            \STATE Mask $\tilde{x}$ synthetically $\to \tilde{x}'$;\ $g \leftarrow \nabla_{\phi}\mathcal{L}_{\text{distill}}(\tilde{x}')|_{\phi_{\text{anchor}}}$
            \STATE $\mathrm{FMI} \leftarrow \frac{1}{p}\sum_j g_j^2 / (F_{\text{anchor},j}+\varepsilon)$;\ push to $\mathcal{B}$ (keep last $M$)
            \IF{$\text{FMI} > \tau_{fmi}$}
                \STATE $\phi, F_{\text{prev}} \leftarrow \phi_{\text{anchor}}, F_{\text{anchor}}$;\ $c \leftarrow 0$;\ $\text{state} \leftarrow \textsc{Active}$;\ $\mathcal{B} \leftarrow \emptyset$
            \ENDIF
        \ENDIF
    \ENDIF
\ENDFOR
\end{algorithmic}
\end{algorithm}

\subsection{Theoretical Properties of FAR}
\label{sec:far-theory}

This section provides an intuitive theoretical justification for the proposed
Fisher-Anchored Restoration (FAR) mechanism. Rather than proving global
optimality, our goal is to explain why the proposed two-state adaptation
strategy avoids unnecessary parameter drift while remaining capable of
re-adapting under distribution shift.

\subsubsection{Assumptions}

The following assumptions are standard in stochastic optimization and
test-time adaptation.

\noindent\textbf{Assumption 1 (Local Smoothness).}
The feature-level distillation objective
$L_{\mathrm{distill}}(\phi)$ is continuously differentiable with
$L$-Lipschitz gradients,

\begin{equation}
\|\nabla L(\phi_1)-\nabla L(\phi_2)\|
\le
L\|\phi_1-\phi_2\|.
\end{equation}

\noindent\textbf{Assumption 2 (Bounded Gradient Variance).}
The stochastic gradients satisfy

\begin{equation}
\mathbb{E}[g_t]
=
\nabla L(\phi_t),
\end{equation}

and

\begin{equation}
\mathbb{E}
\left[
\|g_t-\nabla L(\phi_t)\|^2
\right]
\le
\sigma^2.
\end{equation}

\noindent\textbf{Assumption 3 (Piecewise Stationary Test Stream).}
The test distribution remains stationary between distribution shifts.
This assumption is common in continual test-time adaptation, where
environmental conditions remain approximately constant for finite
intervals before changing.

\subsubsection{Proposition 1: Fisher Stability as a Stationarity Signal.\\}
\label{prop:fisher-stability}
Let $F_t$ denote the online Fisher diagonal defined in Eq.~(12),
\begin{equation}
F_t = \beta_F F_{t-1} + (1-\beta_F)\, g_t^{\odot 2},
\end{equation}
where $g_t$ is the stochastic gradient of $\mathcal{L}_{\text{distill}}$
at step $t$, taken coordinate-wise so that $g_t^{\odot 2}$ denotes
element-wise squaring. Under Assumption~2 (bounded gradient variance)
and Assumption~3 (piecewise stationarity), suppose
\begin{equation}
\mathrm{CI}_t = \frac{\|F_t - F_{t-1}\|_1}{\|F_t\|_1 + \varepsilon} < \tau_{fc}
\end{equation}
holds for $K$ consecutive adaptation steps. Then the second moment of
the stochastic gradient, $\mathbb{E}[g_t^{\odot 2}]$, has stabilized
across those steps, in the sense of Eq.~(9) below. This need not imply
that $\nabla \mathcal{L}(\phi_t) \approx 0$.

\noindent\textbf{Proof Sketch:} From the recursion defining $F_t$,
\begin{equation}
F_t - F_{t-1} = (1-\beta_F)\left(g_t^{\odot 2} - F_{t-1}\right).
\end{equation}
Substituting into the definition of $\mathrm{CI}_t$,
\begin{equation}
\mathrm{CI}_t = \frac{(1-\beta_F)\,\|g_t^{\odot 2} - F_{t-1}\|_1}{\|F_t\|_1+\varepsilon}.
\end{equation}
Since $\beta_F \in (0,1)$ is fixed, $\mathrm{CI}_t < \tau_{fc}$ is
equivalent, up to the constant $(1-\beta_F)$, to
\begin{equation}
\|g_t^{\odot 2} - F_{t-1}\|_1 < \frac{\tau_{fc}}{1-\beta_F}\left(\|F_t\|_1+\varepsilon\right),
\end{equation}
i.e., the current squared gradient agrees with the running Fisher
estimate in aggregate $\ell_1$ mass. Taking expectations under
Assumption~2, which gives $\mathbb{E}[g_t] = \nabla \mathcal{L}(\phi_t)$
and $\mathbb{E}\|g_t - \nabla \mathcal{L}(\phi_t)\|^2 \leq \sigma^2$, the
per-coordinate second moment satisfies
\begin{equation}
\mathbb{E}[(g_{t,j})^2] = (\nabla_j \mathcal{L}(\phi_t))^2 + \mathrm{Var}(g_{t,j}).
\end{equation}
Condition~(5) small for $K$ consecutive steps therefore forces
\begin{equation}
\mathbb{E}[(g_{t,j})^2] \approx \mathbb{E}[(g_{t-1,j})^2] \approx \cdots \approx \mathbb{E}[(g_{t-K,j})^2] \quad \forall j,
\end{equation}
which holds under either of two disjoint cases:
\begin{align}
\text{(a) stationary point:}\quad &\nabla \mathcal{L}(\phi_t) \approx 0, \\
&\mathrm{Var}(g_{t,j}) \approx \sigma_j^2, \\[6pt]
\text{(b) constant residual:}\quad &\nabla \mathcal{L}(\phi_t) \approx c_j \neq 0, \\
&\mathrm{Var}(g_{t,j}) \approx \sigma_j^2 - c_j^2 \quad \text{(unchanged)}.
\end{align}
Equation~(7) alone cannot distinguish case (a) from case (b): both
produce a stable $F_t$. What Eq.~(7) does establish is that the
distribution generating $g_t$, and in particular its second moment, has
stopped changing over the $K$-step window.

\noindent\textbf{Remark (Informal): Consolidation Reduces Unnecessary
Drift.} Given Eq.~(7), consider $T$ additional steps taken while
$\mathrm{CI}_t$ remains below $\tau_{fc}$. Writing $\phi_{t+T} - \phi_t =
-\eta \sum_{i=1}^{T} g_{t+i}$ and applying Assumption~2,
\begin{equation}
\mathbb{E}\left[\|\phi_{t+T}-\phi_t\|^2\right]
= \eta^2 \sum_{i=1}^{T}\mathbb{E}\left[\|g_{t+i}\|^2\right]
= O(T\eta^2 \sigma^2),
\end{equation}
matching Eq.~(5), regardless of whether case (a) or (b) above holds. In
case (a), this accumulation is pure noise with no accompanying decrease
in $\mathcal{L}_{\text{distill}}$; in case (b), useful progress
$-\eta T c$ is being made, but at rate bounded by $c$, the residual
gradient, which Eq.~(7) shows is itself no longer detectably shrinking.
In either case, continued updates trade a fixed, boundable amount of
progress against unbounded, monotonically growing variance as $T$
increases. FAR operationalizes the resulting tradeoff by freezing $\phi$
once Eq.~(2) holds for $K$ consecutive steps, accepting the small
residual-progress cost of case (b) in exchange for eliminating the
variance growth in Eq.~(9) until Eq.~(15) signals that reactivation is
warranted.

\subsubsection{Proposition 2: No-Regret Reactivation\\}
\label{prop:no-regret}
\noindent Assume the test distribution changes from
$P_A$ to $P_B$, and the induced gradient statistics satisfy

\begin{equation}
FMI_t>\tau_{fmi}.
\end{equation}

Then FAR exits the anchored state and resumes gradient updates.
Following reactivation, the optimization dynamics are identical to
continuous TTSD adaptation up to a finite detection delay.

\noindent\textbf{Proof Sketch:}
During the anchored state, the student parameters remain fixed,

\begin{equation}
\phi_t=\phi_{\mathrm{anchor}}.
\end{equation}

For each incoming complete-modality sample, FAR evaluates

\begin{equation}
FMI_t
=
\frac1p
\sum_{j=1}^{p}
\frac{g_{t,j}^2}
{F_{\mathrm{anchor},j}+\epsilon}.
\end{equation}

When the test distribution remains stationary, the incoming gradients
remain aligned with the previously consolidated Fisher geometry,
resulting in relatively small mismatch values that remain below the
reactivation threshold.

Suppose the underlying distribution changes. The gradients then begin
appearing along parameter directions that previously exhibited low
Fisher importance. Consequently,

\begin{equation}
g_{t,j}^2
\gg
F_{\mathrm{anchor},j},
\end{equation}

for a subset of parameters, causing the Fisher mismatch index to exceed
$\tau_{fmi}$.

Once reactivated, FAR performs the same parameter update as standard
TTSD,

\begin{equation}
\phi_{t+1}
=
\phi_t-\eta g_t.
\end{equation}

Therefore, after detecting a distribution shift, FAR follows the same
optimization trajectory as perpetual adaptation. The only difference is
that FAR avoids unnecessary updates during stationary periods, thereby
reducing parameter drift without sacrificing future adaptability.

\subsubsection{Proposition 3: FMI Hard-Sample Robustness.\\}
\label{prop:fmi-robustness}

\noindent Suppose the student LoRA is in the anchored state with stored baseline
$F_{\text{anchor}} \approx F^*(P) := \mathbb{E}_P[g \odot g]$. Consider
two scenarios that produce equally elevated distillation loss:
\begin{equation}
\mathbb{E}_{P'}[\mathcal{L}_{\text{distill}}] = C \cdot
\mathbb{E}_P[\mathcal{L}_{\text{distill}}],
\end{equation}
where scenario (a) is a hard within-distribution sample $x_H \sim P$ with
$\|g_H\|^2 = C \cdot \mathbb{E}_P[\|g\|^2]$, and scenario (b) is a
shifted distribution $P'$ with the same expected loss elevation. Under
Assumptions~2--3, the FMI satisfies
\begin{equation}
\mathbb{E}_P[\mathrm{FMI}(x_H)] \approx 1,
\end{equation}
regardless of $C$, while
\begin{equation}
\mathbb{E}_{P'}[\mathrm{FMI}_t] \neq 1,
\end{equation}
whenever $F^*(P') \neq F^*(P)$. Consequently, the FMI threshold
$\tau_{fmi}$ is triggered selectively by genuine distributional shift and
not by gradient magnitude alone.

\noindent\textbf{Proof Sketch:} By definition of the diagonal Fisher,
$F^*(P)_j = \mathbb{E}_P[g_j^2]$. For a hard sample $x_H \sim P$, the
gradient $g_H$ is drawn from the same distribution $P$ used to construct
$F_{\text{anchor}}$, so $\mathbb{E}_P[g_{H,j}^2] = F^*(P)_j =
F_{\text{anchor},j}$ coordinate-wise — the magnitude $C$ scales the
gradient but is distributed across the same high-importance coordinates
that defined $F_{\text{anchor}}$ in the first place, leaving the ratio
unchanged:
\begin{equation}
\mathbb{E}_P[\mathrm{FMI}(x_H)] = \frac{1}{p}\sum_{j=1}^p
\frac{F_{\text{anchor},j}}{F_{\text{anchor},j}+\varepsilon} \approx 1.
\end{equation}
Under a shifted distribution $P'$, gradients instead concentrate in
coordinates $j \in S$ where $F_{\text{anchor},j}$ was small, so
$g_{t,j}^2 \gg F_{\text{anchor},j}$ for $j \in S$, giving
\begin{equation}
\mathbb{E}_{P'}[\mathrm{FMI}_t] \;\geq\; 1 + \frac{|S|}{p}\cdot
\frac{\delta}{f_{\text{low}}+\varepsilon},
\end{equation}
where $\delta>0$ is the shift magnitude in $S$ and $f_{\text{low}} =
\max_{j\in S} F_{\text{anchor},j}$. As $f_{\text{low}} \to 0$, this bound
diverges, so any finite $\tau_{fmi}$ eventually detects the shift.
Sample difficulty within $P$ (large $C$) therefore leaves FMI near 1,
while structural change in where gradient mass concentrates (a genuine
distributional shift) drives it above threshold.

\section{Additional Results}
\label{sec:additional-results}

\subsubsection{Distilling from Base Teacher (no teacher LoRA).\\}
\label{subsec:distill-notl}

\begin{figure}[t]
\centering
\includegraphics[width=1.0\linewidth]{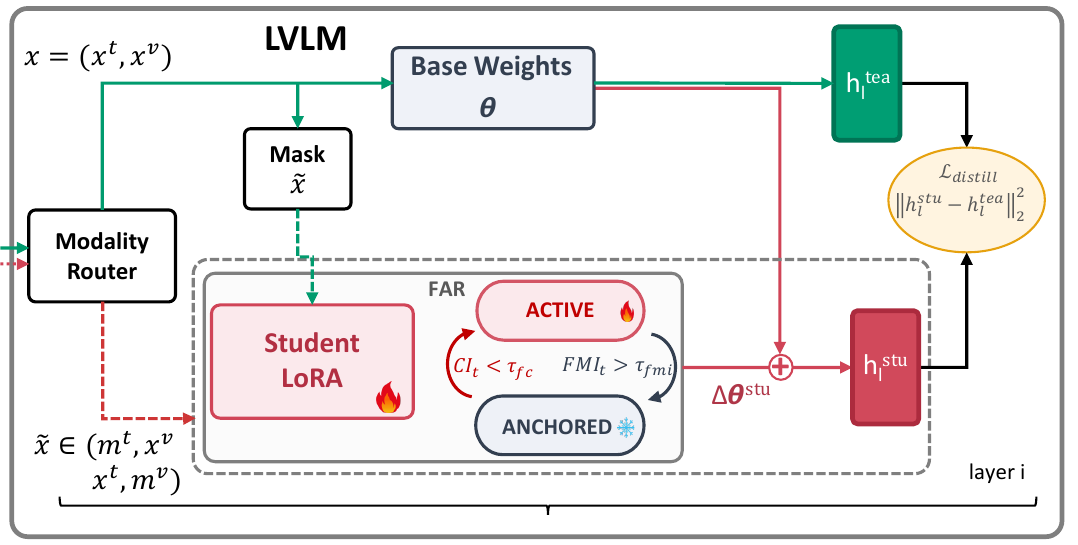}
\caption{\small TTSD-FAR without teacher LoRA. Base weights provide teacher representations.}
\label{fig:ttsd_notl}
\end{figure}

When working with LLMs and LVLMs, it is often realistic to assume that no retraining, either full or parameter-efficient, is performed on the base model. To demonstrate the effectiveness of TTSD under this constraint, we conduct an experiment where distillation is performed directly from the base model weights without any teacher LoRA as shown in Fig. ~\ref{fig:ttsd_notl}. In this setting, teacher representations are extracted from the frozen base model, resulting in the same optimization problem as Eq. 11

Table ~\ref{tab:bah_text_missing-notl} shows the results for the case where the teacher embeddings are obtained directly from the base model weights. Since the same distillation mechanism is applied, the setup is effectively identical except that the teacher representations are extracted from the frozen base model rather than from a Teacher LoRA trained with the objective in Eq. 11. The results therefore follow a similar trend to the case where the teacher embeddings are obtained using the Teacher LoRA objective.

\begin{table}[t]
\centering
\footnotesize
\setlength{\tabcolsep}{1.5pt}
\caption{Comparison of methods under progressively missing text conditions. Distilling from the Base model, i.e., no teacher LoRA. We report WF1 on the BAH dataset as the proportion of unavailable textual input increases.}
\label{tab:bah_text_missing-notl}
\begin{tabular}{lcccccc}
\toprule
\textbf{Method} & \textbf{Full} & \textbf{10\%} & \textbf{20\%} & \textbf{30\%} & \textbf{40\%} & \textbf{50\%} \\
\midrule
No Adaptation & \textbf{0.6341} & 0.5905 & 0.5798 & 0.5710 & 0.5662 & 0.5614 \\
\midrule
TENT & -- & 0.4910 & 0.4840 & 0.4733 & 0.4501 & 0.4453 \\
EATA & -- & 0.5465 & 0.5367 & 0.5290 & 0.5245 & 0.5110 \\
RAG & -- & 0.5770 & 0.5720 & 0.5540 & 0.5448 & 0.5250 \\
Perplexity Gen. & -- & 0.5790 & 0.5695 & 0.5648 & 0.5488 & 0.5430 \\
\textbf{TTSD-FAR (Ours)} & -- & \textbf{0.6185} & \textbf{0.6120} & \textbf{0.5955} & \textbf{0.5910} & \textbf{0.5870} \\
\bottomrule
\end{tabular}
\end{table}

\subsection*{Weaker-Modality Missing Results}
\label{sup-sec:dataset-imp}
\subsubsection*{DFEW - Text Missing at Test-Time - Weaker Modality Missing}

Table~\ref{tab:dfew_text_missing} reports the performance of different methods on the DFEW under progressively increasing levels of missing textual modality. As the proportion of missing text increases, the performance of the model without adaptation gradually degrades.

\begin{table}[h]
\centering
\footnotesize
\setlength{\tabcolsep}{1.5pt}
\caption{Comparison of methods under progressively missing textual modality conditions. F1 on the DFEW dataset as the proportion of missing textual input increases.}
\label{tab:dfew_text_missing}
\begin{tabular}{lc|ccccc}
\toprule
\textbf{Method} & \textbf{Full} & \textbf{10\%} & \textbf{20\%} & \textbf{30\%} & \textbf{40\%} & \textbf{50\%} \\
\midrule
No Adaptation & \textbf{0.5549} & 0.5530 & 0.5490 & 0.5365 & 0.5260 & 0.5235 \\
\midrule
TENT & -- & 0.1655 & 0.1725 & 0.1790 & 0.1935 & 0.1833 \\
EATA & -- & 0.2635 & 0.2674 & 0.2835 & 0.2855 & 0.2943 \\
RAG & -- & 0.5530 & 0.5425 & 0.5265 & 0.5265 & 0.5210 \\
Perplexity Gen. & -- & 0.5535 & 0.5460 & 0.5350 & 0.5300 & 0.5210 \\
\textbf{TTSD-FAR (Ours)} & -- & \textbf{0.5540} & \textbf{0.5510} & \textbf{0.5444} & \textbf{0.5376} & \textbf{0.5278} \\
\bottomrule
\end{tabular}
\end{table}
However, the performance drop due to the missing modality is insignificant compared to the vision missing in the DFEW dataset. This is primarily due to the dataset structure. The DFEW dataset comprises of 16372 short clips, of which 6196 are non-English transcriptions and 4960 clip have no speaker utterance at all. This shows that even the full modality performance (0.5549) is already less reliant on the textual modality. Classical test-time adaptation methods such as TENT \cite{wang2021tent} and EATA \cite{niu2022efficienttesttimemodeladaptation} exhibit severe performance drops, suggesting that entropy-minimization-based adaptation strategies are not well suited for multimodal missing-modality scenarios. In contrast, retrieval-based augmentation (RAG) and generation-based approaches maintain performance because the observed modality is stronger in this case. Our proposed \textbf{TTSD} method consistently achieves the best performance under moderate to severe missing-text conditions (20\%–50\%), demonstrating improved robustness and stability when textual inputs are partially unavailable. These results indicate that TTSD effectively mitigates the negative impact of missing textual modality while preserving strong performance as modality degradation increases.

\subsubsection*{BAH - Vision Missing at Test-Time- Weaker Modality Missing}

Table~\ref{tab:bah_vision_missing} reports results on BAH under vision missing conditions. The vision drop for the BAH dataset results in negligible performance drops even at severe missing percentages (50\%). The 10\% has no effect on the performance.   
TTSD is able to improve over the existing TTA methods in both weaker and stronger missing modality conditions. 
\begin{table}[h]
\centering
\footnotesize
\setlength{\tabcolsep}{1.5pt}
\caption{Comparison of methods under progressively missing visual modality. We report Avg-F1 on the BAH dataset as the proportion of unavailable visual input increases.}
\label{tab:bah_vision_missing}
\begin{tabular}{lc|ccccc}
\toprule
\textbf{Method} & \textbf{Full} & \textbf{10\%} & \textbf{20\%} & \textbf{30\%} & \textbf{40\%} & \textbf{50\%} \\
\midrule
No Adaptation & \textbf{0.7142} & 0.7142 & 0.7095 & 0.7046 & 0.6950 & 0.6923 \\
\midrule
TENT & -- & 0.4834 & 0.4846 & 0.4790 & 0.4645 & 0.4754 \\
EATA& -- & 0.5655 & 0.5428 & 0.5165 & 0.5095 & 0.5052 \\
RAG & -- & 0.7100 & 0.7055 & 0.7059 & 0.7110 & 0.6975 \\
\textbf{TTSD-FAR (Ours)} & -- & \textbf{0.7142} & \textbf{0.7125} & \textbf{0.7120} & \textbf{0.7125} & \textbf{0.7125} \\
\bottomrule
\end{tabular}
\end{table}

As shown in Table \ref{tab:prompt_variations}, the textual input is critical for Ambivalence detection. This explains why the text-missing case for the BAH dataset results in a much higher performance drop than vision-missing, as shown in Table \ref{tab:bah_vision_missing}. Another important observation is that in the vision missing case (Table \ref{tab:bah_vision_missing}) the RAG-based method is also able to improve, unlike the text missing case. This is because in the vision missing case, the observed modality (text) is stronger and is able to effectively retrieve relevant samples. Whereas, in the text missing case, the observed modality (vision) was not discriminant enough for effective retrieval.

\section{Additional Ablations:}
\label{sec:additional-ablations}

\noindent\subsubsection{LoRA Rank Sensitivity.} 
\label{sec:lora-rank-sensitivity}
We investigate the effect of LoRA rank on TTSD's robustness to modality-incomplete inputs. As expected, performance degrades monotonically with increasing missing rates across all ranks; yet, the degradation pattern reveals a clear sensitivity to rank capacity. Lower-rank adaptation ($r$=4) suffers disproportionately under higher missing rates, exhibiting steeper and less stable declines compared to higher rank, suggesting that insufficient parameter capacity limits the model's ability to compensate for missing modalities. Ranks $r$=8 and $r$=16 remain consistently close throughout, with only marginal gains from doubling the rank, indicating diminishing returns beyond $r$=8. These results suggest that $r$=8 strikes the best balance between adaptability, robustness, and computational complexity. The ablation shows that LoRA rank is a meaningful factor in missing-modality robustness.

\begin{figure}[h]
  \centering
  \includegraphics[height=2.7cm]{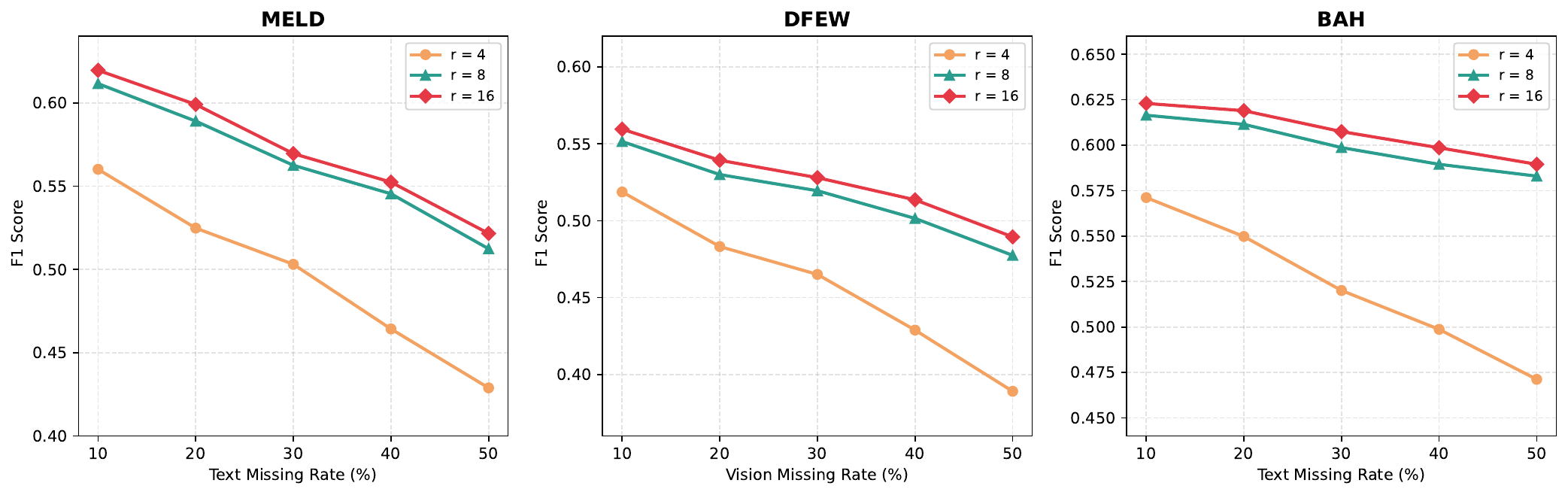}
  %\caption{Effect of LoRA rank on distillation performance for the three datasets.} ALEKOE
  \caption{Impact of LoRA rank on distillation performance under varying missing-modality rates across three datasets.}
  \label{fig:lora_ablation}
\end{figure}

\subsubsection*{Effectivity of TTSD on Small Video Language Model (SVLM):}
\label{sec:svlm}
To further show the effectiveness of the proposed TTSD method. We perform the experiment on a smaller video language model, MobileVideoGPT-0.5B \cite{shaker2025mobilevideogptfastaccuratevideo}.

\begin{table*}[h]
\centering
% \footnotesize
\caption{F1 Score on the BAH dataset under progressively missing textual input for two models: Video-LLaVA-7B and MobileVideoGPT-0.5B.}
\label{tab:dfew_text_missing_models}
\begin{tabular}{l|c|c|ccccc}
\toprule
\textbf{Model} & \textbf{Method} & \textbf{Complete} & \textbf{10\%} & \textbf{20\%} & \textbf{30\%} & \textbf{40\%} & \textbf{50\%} \\
\midrule

\multirow{2}{*}{Video-LLaVA-7B} 
& No Adaptation & \multirow{2}{*}{\textbf{0.6341}} & 0.5905 & 0.5798 & 0.5710 & 0.5662 & 0.5614 \\
& TTSD-FAR (Ours) &  & \textbf{0.6185} & \textbf{0.6120} & \textbf{0.5955} & \textbf{0.5910} & \textbf{0.5870}\\

\midrule

\multirow{2}{*}{MobileVideoGPT-0.5B} 
& No Adaptation & \multirow{2}{*}{0.4889} & 0.4521 & 0.4413 & 0.4310 & 0.4218 & 0.4125 \\
& TTSD-FAR (Ours) &  & \textbf{0.4685} & \textbf{0.4595} & \textbf{0.4510} & \textbf{0.4428} & \textbf{0.4385} \\

\bottomrule
\end{tabular}
\end{table*}

For Video-LLaVA-7B, the baseline model without adaptation shows a gradual degradation in performance as the proportion of missing textual input increases. Applying TTSD consistently improves the results across all missing-modality levels, demonstrating that the proposed method effectively recovers useful representations when textual information becomes partially unavailable. A similar trend is observed for the smaller MobileVideoGPT-0.5B model. Although its overall performance is lower due to its reduced model capacity, TTSD still provides consistent improvements over the non-adapted baseline under all missing-modality conditions. These results indicate that TTSD is not restricted to a specific LVLM architecture and can improve robustness for both large and lightweight video-language models.

\subsection{FAR Hyperparameters Sensitivity}
\label{sec:far-hyperparam-sensitivity}
\subsubsection{Effect of the FMI Threshold:}
\label{fig:ttsd_fmiplot}
\begin{figure}[h]
\centering
\includegraphics[width=0.98\linewidth]{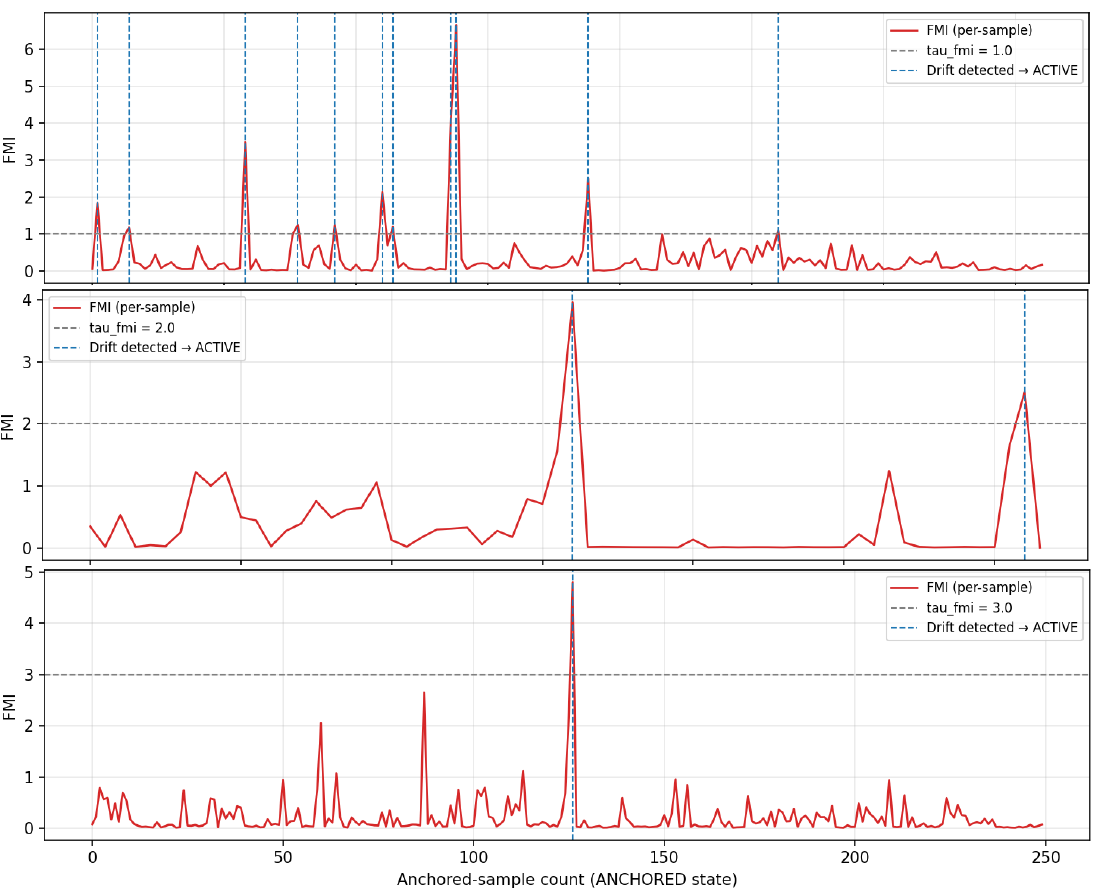}
\caption{\small Effect of $\tau_{fmi}$ on reactivation events. The three panels show the FMI trace at $\tau_{fmi}$ 1.0 (top), 2.0 (middle) and, 3.0 (bottom).}
\label{fig:ttsd_fmiplot}
\end{figure}

Figure~\ref{fig:ttsd_fmiplot} illustrates the sensitivity of FAR's reactivation criterion to the choice of $\tau_{fmi}$. Each panel plots the FMI trace computed during the ANCHORED state, with the horizontal dashed line indicating $\tau_{fmi}$ and vertical dashed lines marking each detected reactivation event. At $\tau_{fmi}=1.0$ (top), the threshold is low enough that it is crossed both by the largest, genuine spikes and by several smaller fluctuations in the 1.0--1.5 range, yielding nine separate reactivations. This demonstrates the failure mode of an overly permissive threshold, where the mechanism reacts to noise as frequently as to true distributional shift. At $\tau_{fmi}=2.0$ (middle), corresponding to the value used in our reported experiments, the threshold clears the minor fluctuations, while still detecting the two clearly dominant spikes, resulting in exactly two reactivations. This setting represents the balance adopted for the main results. At $\tau_{fmi}=3.0$ (bottom), the threshold is raised further such that even a second sizeable spike is no longer detected, leaving only the single largest event to trigger reactivation. This illustrates the opposite failure mode: an excessively conservative threshold trades false positives for false negatives, risking missed detections of genuine distributional shift.

\subsubsection{FAR State Dynamics over the Test Stream}
\label{sec:far-state-dynamics}

\begin{figure}[h]
\centering
\includegraphics[width=0.98\linewidth]{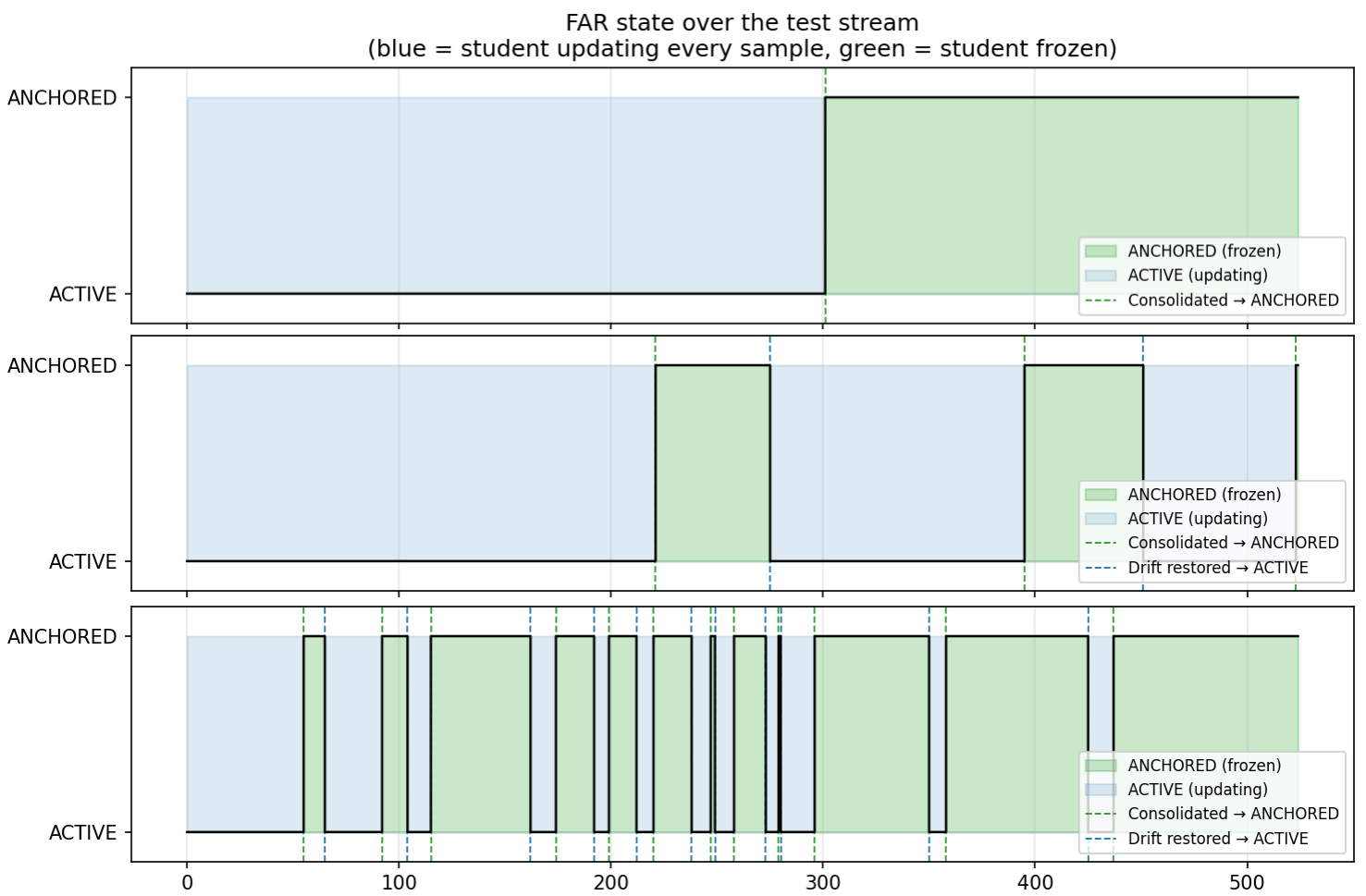}
\caption{\small Timeline of FAR states (\textsc{active}-\textsc{anchored}) across the test stream.}
\label{fig:ttsd_farstates_plot}
\end{figure}

Figure~\ref{fig:ttsd_farstates_plot} shows the ACTIVE/ANCHORED state trajectory of FAR across the test stream under three hyperparameter regimes. The middle panel corresponds to the values reported in the paper and supplementary ($\beta_F=0.99$, $\tau_{fc}=0.02$, $K=5$, $\tau_{fmi}=2.0$). The student remains ACTIVE for roughly the first 40\% of the stream before the
Consolidation Index stays below $\tau_{fc}$ for $K$ consecutive samples, triggering consolidation into ANCHORED. It remains frozen until a distributional shift shortly after, around the mid-point of the stream, raises the FMI above $\tau_{fmi}$, triggering reactivation. A second cycle follows later in the stream, consolidating and then reactivating again, after which the student remains active for the remainder until the very end, where it transitions to the Anchored state. This produces
two well-separated reactivation cycles.

The top panel shows a single consolidation slightly past the mid-point of the stream, with no reactivation for the remainder. This is consistent with a substantially higher $\tau_{fmi} = 5$, under which later shifts in the stream no longer raise the FMI enough to cross the threshold, leaving the student permanently frozen once anchored. 
The bottom panel uses a relaxed consolidation requirement with a lower reactivation threshold of $\tau_{fmi}=0.5$. Consolidation fires as soon as CI drops below $\tau_{fc}$, and the low $\tau_{fmi}$ causes ordinary gradient noise while anchored to be sufficient to trigger reactivation. These settings produce rapid, near-continuous alternation between ACTIVE and ANCHORED throughout most of the stream, resulting in the student being in the ANCHORED state for roughly 80\% of the test stream. These two extremes motivate the reported configuration of $K=5$ and $\tau_{fmi}=2.0$, which yields the stable, interpretable cycles shown in the middle panel rather than premature permanent freezing or unstable oscillation.

\subsection{Loss-Based Stopping vs. FAR}
Table~\ref{tab:loss-stop-ablation} compares TTSD-FAR against two ablated variants on the BAH dataset (with settings similar to Tab.~\ref{tab:bah_text_missing-notl}) under progressively missing textual input. i) plain TTSD, which adapts on every complete-modality sample without any stopping criterion, and ii) TTSD-LS, which instead halts adaptation once the raw distillation loss plateaus with a patience of 5, after which the student remains frozen for the rest of the stream. TTSD-LS underperforms not only TTSD-FAR but also plain, unbounded TTSD at every missing ratio. This confirms the motivation given in Section 3.4: the raw distillation loss is an unreliable stopping signal, since it can plateau due to sample difficulty rather than genuine parameter convergence. Once frozen, TTSD-LS has no mechanism to detect that the test stream has since shifted and never resumes adaptation. It accumulates the disadvantages of stopping without any of the drift-awareness that FAR provides. In contrast, TTSD-FAR consistently outperforms both plain TTSD and TTSD-LS across all missing ratios, indicating that gating adaptation on the Fisher information geometry rather than on the loss itself is necessary to obtain the benefits of stopping without sacrificing the ability to
resume adaptation when genuine distributional shift occurs.

\begin{table}[h]
\centering
\footnotesize
\setlength{\tabcolsep}{1.5pt}
\caption{Ablation comparing TTSD-FAR against plain TTSD (no stopping criterion) and TTSD-LS (loss-plateau stopping criterion) under progressively missing text conditions. We report WF1 on the BAH dataset as the proportion of unavailable textual input increases.}
\label{tab:loss-stop-ablation}
\begin{tabular}{lcccccc}
\toprule
Method & Full & 10\% & 20\% & 30\% & 40\% & 50\% \\
\midrule
No-Adaptation   & 0.6341 & 0.5905 & 0.5798 & 0.5710 & 0.5662 & 0.5614 \\
\midrule
TTSD            & --     & 0.6065 & 0.6039 & 0.5887 & 0.5816 & 0.5728 \\
TTSD-LS         & --     & 0.5910 & 0.5820 & 0.5730 & 0.5645 & 0.5560 \\
\textbf{TTSD-FAR (Ours)} & -- & \textbf{0.6185} & \textbf{0.6120} & \textbf{0.5955} & \textbf{0.5910} & \textbf{0.5870} \\
\bottomrule
\end{tabular}
\end{table}

\section{Datasets and Implementation Details}
\label{sec:datasets-impl}

\subsection*{Datasets}
\label{sec:datasets}
\noindent \textbf{MELD (Multimodal EmotionLines Dataset):} MELD is a conversational emotion recognition dataset that extends the original EmotionLines corpus by incorporating synchronized audio, visual, and textual modalities  \cite{meld_ds}. It contains approximately 1,433 multi-party dialogues and over 13,000 utterances extracted from the \textit{Friends} TV series, each labeled with one of seven discrete emotions (Anger, Disgust, Sadness, Joy, Neutral, Surprise, Fear) and sentiment labels. The multimodal nature and conversational context make MELD a challenging benchmark for models that must integrate semantic, acoustic, and facial cues to infer emotion.

\noindent \textbf{DFEW (Dynamic Facial Expression in the Wild):} DFEW is a large-scale dynamic facial expression dataset collected from more than 1,500 movies and comprising over 16,000 short video clips depicting naturalistic expressions captured under unconstrained conditions \cite{dfew_ds}. Unlike posed or laboratory-controlled datasets, DFEW includes significant variations in pose, illumination, occlusion, and actor demographics, with each clip annotated for one of seven basic emotions. This dataset serves as a benchmark for dynamic facial expression recognition (DFER) in the wild, emphasizing robustness to real-world noise and temporal dynamics.

\noindent \textbf{BAH (Behavioral Ambivalence/Hesitancy Dataset):} The BAH dataset is a recently introduced benchmark for recognizing ambivalence and hesitancy (A/H) in video recordings designed for behaviour change studies \cite{gonzalez-26-bah}. It contains 1,118 videos recorded from 224 participants across diverse demographics, captured while participants responded to stimuli intended to elicit ambivalence or hesitancy. Frame- and video-level annotations highlight segments containing A/H cues, and the dataset also provides aligned face crops, audio transcripts with timestamps, and participant metadata. BAH’s focus on subtle and conflicting emotional states presents a distinct challenge compared to conventional discrete emotion recognition tasks, highlighting the need for models that can capture fine-grained behavioral cues across modalities and are able to perform well even under missing conditions. The dataset comes with a predefined train-test split. We present the results on the official test set. The performance measure reported is the F1 score.
\begin{table*}[h]
% \label{tab:prompt_summary}
\centering
\footnotesize
\renewcommand{\arraystretch}{1.15}
\caption{Summary of prompt variations for zero-shot inference with corresponding video-level F1 scores.}
\label{tab:prompt_variations}
\begin{tabularx}{\linewidth}{>{\centering\arraybackslash}m{2.2cm}|>{\raggedright\arraybackslash}X|>{\centering\arraybackslash}m{1.2cm}}
\hline
\textbf{Prompt Type} & \textbf{Prompt} & \textbf{Avg F1} \\
\hline
Simple &
Classify the emotion in the video as either \textit{Non-Ambivalent} or
\textit{Ambivalent}. Respond with only one word.
& 0.2827 \\
\hline
Definition 1 &
Definition: Ambivalence is the state of having contradictory or conflicting
feelings or attitudes towards something or someone simultaneously.
Classify the emotion in the video as either \textit{Non-Ambivalent}
or \textit{Ambivalent}. Respond with only one word.
& 0.3326 \\
\hline
Definition 2 &
Definition: Ambivalence and hesitancy is understood as the simultaneous
experience of desires for change and against change.
Classify the emotion in the video as either \textit{Non-Ambivalent}
or \textit{Ambivalent}. Respond with only one word.
& 0.3772 \\
\hline
Transcript + Def 1 &
Video transcript: \textcolor{red}{\{transcript\}}.
Definition: Ambivalence is the state of having contradictory or
conflicting feelings or attitudes towards something or someone
simultaneously. Classify the emotion in the video as either
\textit{Non-Ambivalent} or \textit{Ambivalent}.
Respond with only one word.
& \textbf{0.6341} \\
\hline
Transcript + Def 2 &
Video transcript: \textcolor{red}{\{transcript\}}.
Definition: Ambivalence and hesitancy are understood as the
simultaneous experience of desires for change and against change.
Classify the emotion in the video as either
\textit{Non-Ambivalent} or \textit{Ambivalent}.
Respond with only one word.
& 0.3945 \\
\hline
\end{tabularx}
\end{table*}

\subsection*{Implementation Details}
\label{sec:impl-details}
\subsubsection{Missing modality Simulation:}
\begin{figure}[h]
  \centering
  \includegraphics[height=4.2cm]{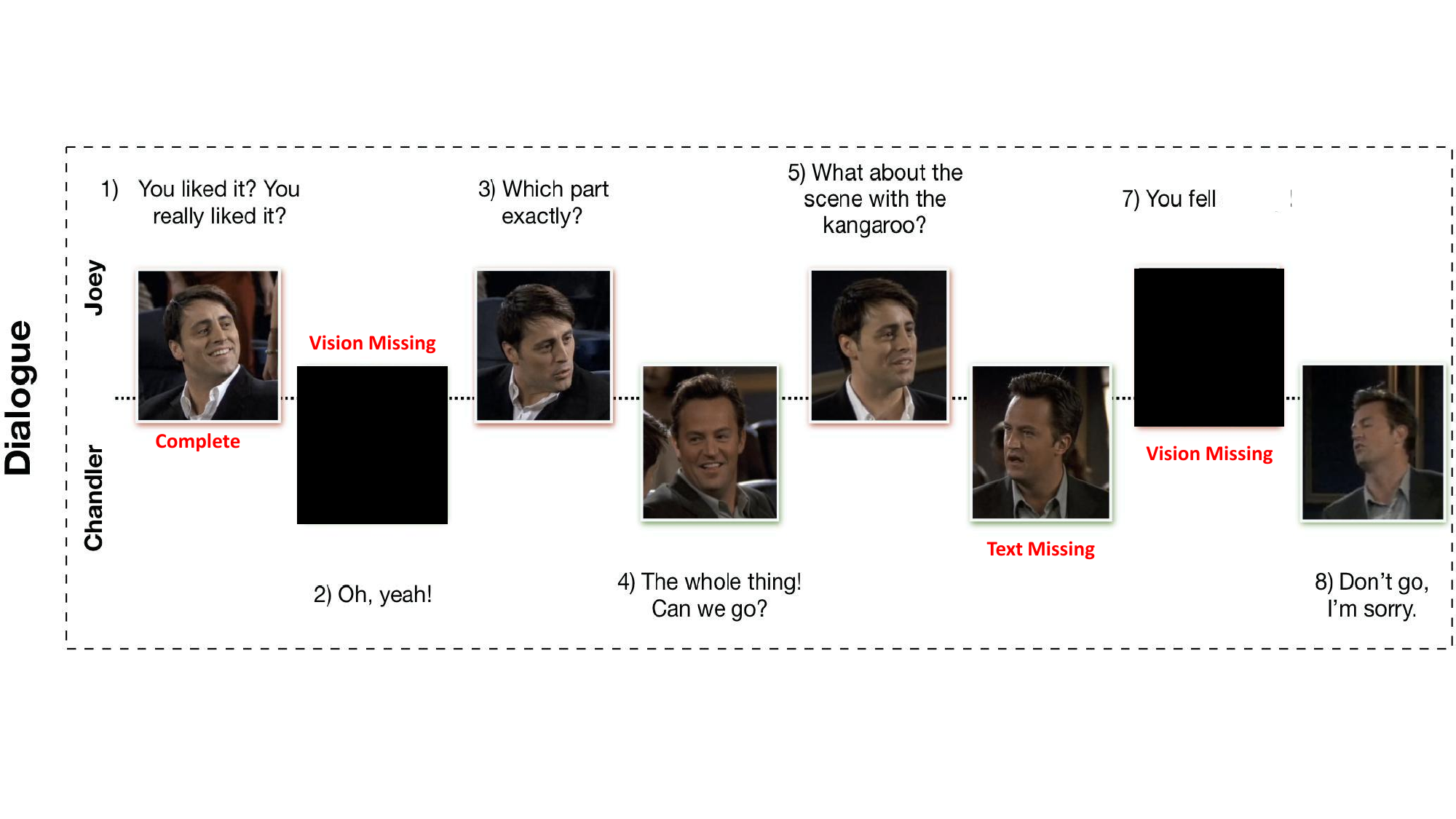}
  \caption{Illustration of the missing modality  
}
  \label{fig:missing_img}
\end{figure}

This section provides the details of the missing modality simulation for both vision and text missing conditions. The missing modality problem is a vast paradigm, and it is important to set the scope of the experimentation to effectively validate the method. 
The datasets used for the validation of the TTSD-FAR method all contain utterance-level annotations. For the text-missing scenario, the speaker utterance in the prompt is replaced by an empty string, and for the vision-missing scenario, zero-imputed frames tensor is fed into the model. Figure \ref{fig:missing_img} shows the vision-missing and text-missing scenarios for the MELD dataset.

\noindent \textbf{MELD Dataset:} We evaluate on the MELD test split using utterance-level metadata providing dialogue ID, speaker, transcription, and 7-class emotion labels (neutral, surprise, fear, sadness, joy, disgust, anger). Visual inputs are constructed from pre-extracted cropped and aligned facial frames, resized to $224\times224$ and uniformly sampled to $T=8$, frames per utterance. Textual inputs are the raw utterance transcriptions from the metadata. Missing modality scenarios are simulated by replacing the utterance string with an empty string for text-missing conditions, controlled by a seeded random state (seed 42) for reproducibility. The model is prompted in instruction-format with the video and utterance as inputs, and asked to classify the emotion from the seven candidate labels. Student LoRA is adapted using the Adam optimizer with a learning rate $10^{-4}$ with a batch size of 1, where gradient accumulation over 8 steps is equivalent to performing a single effective update per sample, updating only the student LoRA parameters $ r=8$ , $\alpha=16$, and dropout 0.3. 
\noindent \textbf{DFEW Dataset:} Since Video-LLaVA-7B is used in all experiments, the data loading and prompting structure remains the same as MELD. Student LoRA is adapted using the Adam optimizer with a learning rate $10^{-4}$ with a batch size of 1, where gradient accumulation over 8 steps is equivalent to performing a single effective update per sample, updating only the student LoRA parameters $ r=8$ , $\alpha=16$, and dropout 0.3.
All experiments use float16 precision. The results are reported on the Official Fold-1 test set. 

\noindent \textbf{BAH Dataset:} Similar to DFEW and MELD datasets, we perform the experiments for the BAH dataset using the VideoLlava model. However, the task of Ambivalence/Hesitancy detection is more complex than basic emotion recognition. Following Gonzales \etal \cite{gonzalez-26-bah}, we use the prompting structure shown in Table ~\ref{tab:prompt_variations} to get the best results.  Table ~\ref{tab:prompt_variations} shows high reliance on the textual input. Adding an accurate description of the task improves performance, and the best results are achieved with the definition and speaker utterance.

Student LoRA is adapted using the Adam optimizer with a learning rate $10^{-4}$ with a batch size of 1, where gradient accumulation over 8 steps is equivalent to performing a single effective update per sample, updating only the student LoRA parameters $r=8$, $\alpha=16$, and dropout 0.3.
All experiments use float16 precision. The results are reported on the official test set. 
\\
\noindent \textbf{Compute details:} For all experiments, the student is being aligned to the teacher's hidden representations, not its output distribution. We therefore fix generation to greedy decoding (do\_sample=False) so that final predictions are deterministic given a fixed adapter state. Experiments were run on a machine with 4$\times$ \textbf{\textit{NVIDIA A100-SXM4-40GB GPUs}} and CUDA version 12.8. The exact libraries and corresponding versions are included in the code supplement.

\subsubsection{FAR Parameters:}
$\beta_F$, $\tau_{fc}$, $K$, and $\tau_{fmi}$ together control the ACTIVE/ANCHORED state machine. $\beta_F$ sets the EMA decay for the Fisher diagonal. $\tau_{fc}$ and $K$ jointly gate consolidation: CI must stay below $\tau_{fc}$ for $K$ consecutive adapted samples before the student is frozen. $\tau_{fmi}$ gates reactivation, triggering once FMI exceeds this threshold. $M$ and $\varepsilon$ play a minor role, affecting only reporting and numerical stability rather than the transition logic itself.

\begin{table}[h]
\centering
\caption{FAR hyperparameters}
\renewcommand{\arraystretch}{1.3}
\begin{tabular}{c|c|p{4.8cm}}
\hline
\textbf{Parameter} & \textbf{Value} & \textbf{Definition} \\
\hline
$\beta_F$      & 0.99      & EMA decay parameter for the online Fisher diagonal \\
\hline
$\tau_{fc}$    & 0.02      & Consolidation Index threshold \\
\hline
$K$            & 5         & No. of consecutive adaptation steps required \\
\hline
$\tau_{fmi}$   & 2.0       & Fisher Mismatch Index threshold \\
\hline
$\varepsilon$  & $10^{-8}$ & Numerical stability constant \\
\hline
\end{tabular}

\label{tab:far-params-brief}
\end{table}

\bibliography{aaai2027}


\end{document}